\documentclass{article}

\usepackage{microtype}
\usepackage{graphicx}
\usepackage{subfigure}
\usepackage{booktabs} 
\usepackage{hyperref}

\usepackage[accepted]{icml2024}
\usepackage{amsmath}
\usepackage{amssymb}
\usepackage{mathtools}
\usepackage{amsthm}
\usepackage[capitalize,noabbrev]{cleveref}

\usepackage{url}            
\usepackage{amsfonts}       
\usepackage{xcolor}         
\usepackage{enumitem}
\usepackage{array}
\usepackage{float}
\usepackage{alphalph}
\usepackage{bbm}
\usepackage{graphicx} 
\usepackage{multicol}
\usepackage{tikz}
\usepackage{multirow}

\theoremstyle{plain}
\newtheorem{theorem}{Theorem}[section]

\newtheorem{lemma}[theorem]{Lemma}
\newtheorem{corollary}[theorem]{Corollary}
\theoremstyle{definition}

\theoremstyle{remark}

\usepackage[textsize=tiny]{todonotes}

\icmltitlerunning{Logistic Variational Bayes Revisited}

\newcommand{\E}{\mathbb{E}}             
\newcommand{\R}{\mathbb{R}}             
\newcommand{\D}{\mathcal{D}}            
\newcommand{\Q}{\mathcal{Q}}            
\newcommand{\F}{\mathcal{F}}          
\newcommand{\KL}{D_\text{KL}}             
\newcommand{\GP}{\text{GP}}             

\newcommand{\sigmoid}{s}

\newcommand{\argmin}{{\arg\!\min}}      

\DeclareMathOperator{\diag}{diag}

 \newcommand{\thickline}[1]{\raisebox{2pt}{%
    \protect\tikz{ \protect\draw[-,#1,solid,line width = 1.2pt](0,0) -- (5mm,0);}}}
 \newcommand{\dashedline}[1]{\raisebox{2pt}{%
    \protect\tikz{ \protect\draw[-,#1,dashed,line width = 0.8pt](0,0) -- (5mm,0);}}}
\newcommand{\pointFilled}[1]{\raisebox{1.2pt}{%
   \protect\tikz{ \protect\draw[-,#1,fill=#1](0,0) circle (1pt);}}}

\definecolor{tabgreen}{HTML}{2ca02c}
\definecolor{tabblue}{HTML}{1f77b4}
\definecolor{taborange}{HTML}{ff7f0e}
\definecolor{magenta}{HTML}{ff00ff}
\usepackage{lipsum}

\begin{document}

\twocolumn[
\icmltitle{Logistic Variational Bayes Revisited}

\icmlsetsymbol{equal}{*}

\begin{icmlauthorlist}
\icmlauthor{Michael Komodromos}{yyy}
\icmlauthor{Marina Evangelou}{yyy}
\icmlauthor{Sarah Filippi}{yyy}

\end{icmlauthorlist}

\icmlaffiliation{yyy}{Department of Mathematics, Imperial College London, United Kingdom}

\icmlcorrespondingauthor{Michael Komodromos}{mk1019@ic.ac.uk}

\icmlkeywords{Machine Learning, ICML}

\vskip 0.3in
]

\printAffiliationsAndNotice{}  

\begin{abstract}
  Variational logistic regression is a popular method for approximate Bayesian inference seeing wide-spread use in many areas of machine learning including: Bayesian optimization, reinforcement learning and multi-instance learning to name a few. However, due to the intractability of the Evidence Lower Bound, authors have turned to the use of Monte Carlo, quadrature or bounds to perform inference, methods which are costly or give poor approximations to the true posterior. 
  In this paper we introduce a new bound for the expectation of softplus function and subsequently show how this can be applied to variational logistic regression and Gaussian process classification. Unlike other bounds, our proposal does not rely on extending the variational family, or introducing additional parameters to ensure the bound is tight. In fact, we show that this bound is tighter than the state-of-the-art, and that the resulting variational posterior achieves state-of-the-art performance, whilst being significantly faster to compute than Monte-Carlo methods.
\end{abstract}

\section{Introduction} \label{sec:introduction}

Logistic regression involves modelling the probability of a binary response $y_i \in \{0, 1\}$ given a set of covariates $x_i \in \mathbb{R}^p$ for $i=1, \dots, n$. Formally, 
\begin{equation*}
    y_i \sim \text{Bernoulli}(p_i), \quad 
    p_i = \sigmoid ( f(x_i) ) = \frac{1}{1 + \exp(- f(x_i))}
\end{equation*}
where $p_i$ is the probability of observing $y_i=1$, $f : \R^p \rightarrow \R$ is the unknown  model function, and $s(\cdot)$ is the sigmoid function. 

In the context of Bayesian inference the goal is to compute the posterior distribution of $f$ given the data $\D = \{ (y_i, x_i) \}_{i=1}^n$. In simple settings, such as when $f$ takes the parametric form, $f(x) = x^\top \beta$, where $\beta \in \R^p$ is the coefficient vector, methods such as Markov Chain Monte Carlo (MCMC) can be used to sample from the posterior distribution. However, for large $p$ MCMC is known to perform poorly. Alternatively, when $f$ takes a non-parametric form, as in Logistic Gaussian Process (GP) Classification, MCMC does not scale well with $n$ \citep{Kuss2005, RasmussenW06}.

To address these limitations practitioners have turned to Variational Inference (VI), which 
seeks to approximate the posterior distribution with an element from a family of distributions known as the variational family \citep{Blei2017,Zhang2019a}. Formally, this involves computing 
an approximate variational posterior, given by the minimizer of the Kullback-Leibler (KL) divergence between the posterior, $\pi(\cdot | \D)$, and a distribution within the variational family, $\Q'$,
\begin{equation} \label{eq:kl}
    \tilde{q}(\cdot) = \underset{q(\cdot) \in \Q'}{\argmin}\ \KL (q(\cdot) \ || \ \pi(\cdot | \D)).
\end{equation}
Typically, the variational family, $\Q'$, is chosen to be a family of Gaussian distributions,
\begin{equation}
    \Q' = \left\{ N_d(\mu, \Sigma) \ : \mu \in \R^d, \Sigma \in \mathbb{S}_+^{d} \right\},
\end{equation}
whereupon restricting $\Sigma = \diag(\sigma_1^2, \dots, \sigma_d^2)$, gives rise to a mean-field Gaussian variational family, which we denote by $\Q$. This choice is typically made for computational convenience and often leads to tractable optimization problems \cite{bishop2007}. 

In practice however, the KL divergence in \eqref{eq:kl} is intractable
and cannot be optimized directly, and so the Evidence Lower Bound (ELBO),
\begin{equation} \label{eq:elbo}
    \text{ELBO}(q(\cdot)) = 
    \E_{q(\cdot) } \left[ 
        \ell(\D | \cdot)
    \right]
    - \KL(q(\cdot) \| p(\cdot) ) 
\end{equation}
is maximized instead, where 
$  
    \ell(\D | \cdot) =
     \log \prod_{i=1}^n p(y_i | x_i, \cdot)
$ 
is the log-likelihood function and $p(\cdot)$ is the prior, which we set to a Gaussian with zero mean vector and identity covariance throughout. 

\begin{figure*}[ht]
    \centering
    \centering
    \makebox[\textwidth][c]{ 
    \includegraphics[width=1.00\textwidth]{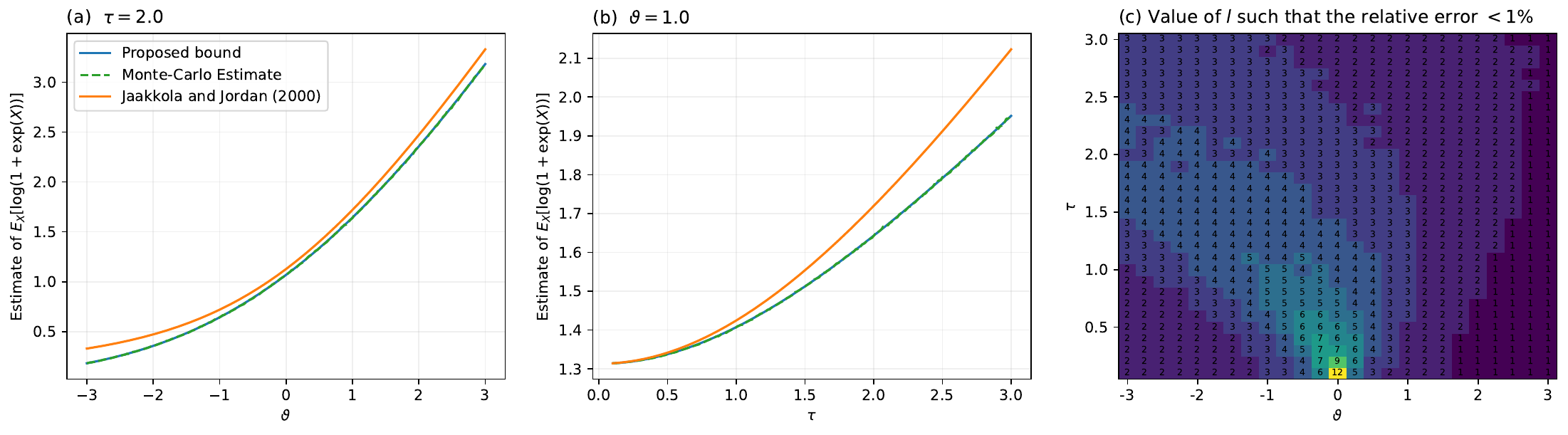}     
    }
    \vspace{-2em}
    \caption{\textit{Error of bounds.} Comparison of \citeauthor{Jakkola97} (\citeyear{Jakkola97}) bound (\thickline{taborange}), proposed bound (\thickline{tabblue}) with $l=10$, and Monte Carlo estimate (\dashedline{tabgreen}) for (a) $\tau=2.0$ and $\vartheta \in [-3, 3]$, (b) $\vartheta = 1.0$ and $\tau \in [0.1, 3.0]$. (c) The number terms ($l$) needed such that the relative error is below 1\%.}
    \label{fig:error_of_bound}
    \vspace{-0.2em}
\end{figure*}

In the context of variational logistic regression there is a further limitation wherein the expected value of the log-likelihood is intractable. This arises through the need to compute the expectation $\E_{X} [ \log(1 + \exp(X))]$ for some $X$. This limitation has led to numerous methods which seek to make this expectation tractable \citep{Depraetere2017a}. Most notably is the seminal work of \citeauthor{Jakkola97} (\citeyear{Jakkola97}), which introduced the quadratic bound,
\begin{equation} \label{eq:jj_bound}
\begin{aligned}
    \log( 1 + \exp(x&))
    =  - \log \sigmoid(-x) \\
    \leq & 
	- \log \sigmoid(t)
	    + \frac{x+t}{2}
	    +
	    \frac{a(t)}{2} (x^2 - t^2)
\end{aligned}
\end{equation}
where $a(t) = \frac{\sigmoid (t) - 1/2}{t}$ and $t$ is a variational parameter that must be optimized to ensure the bound is tight. The bound introduced by \citeauthor{Jakkola97} (\citeyear{Jakkola97}) is tractable under the expectation with respect to $q \in \Q'$, meaning an analytic form of the ELBO in \eqref{eq:elbo} can be derived and optimized with respect to the variational parameters. 

As a result, this bound has seen widespread use in the machine learning community, with applications ranging from Thomson sampling for logistic contextual bandits \cite{Xin2021}, high-dimensional variational logistic regression \cite{Ray2020, komodromos2023} and multi-instance learning with Gaussian processes \cite{Haubmann2017} to name a few.

More recently, a connection between \eqref{eq:jj_bound} and conditionally conjugate Polya-Gamma (PG) logistic regression has been established \cite{Polson2013,Durante2019}. Notably, \citeauthor{Durante2019} \yrcite{Durante2019} showed that the ELBO maximized by \citeauthor{Jakkola97} \yrcite{Jakkola97} is equivalent to 
the ELBO under an extended Polya-Gamma variational family, $\Q' \times \{ \prod_{i=1}^n \text{PG}(1, t_i) \}$ where $\text{PG}(1, t_i)$ is the Polya-Gamma distribution. In turn, this means that \eqref{eq:jj_bound} has a clear probabilistic interpretation, and in fact, is equivalent to optimizing a genuine ELBO rather than a surrogate bound of the ELBO. However, this equivalence highlights the use of a mean-field extension, which in general is known to underestimate the posterior variance \citep{Giordano2018, Durante2019}.

Nevertheless, the Polya-Gamma formulation has been applied to both logistic regression \citep{Durante2019} and Logistic Gaussian Processes \citep{Wenzel2017}. However, fundamentally these methods optimize the same objective as in \citet{Jakkola97}, meaning methods such as those of \citeauthor{Wenzel2017} (\citeyear{Wenzel2017}) coincide with earlier works, e.g. those seen in \citeauthor{Gibbs2000} (\citeyear{Gibbs2000}). 

Beyond these bounds authors have also considered the use of alternative link functions to make computations tractable. For example, via the probit link function which leads to an analytically tractable ELBO \cite{Wang2021}. However, this approach is not without its limitations, as the probit link function is known to be sensitive to outliers \citep{bishop2007}. 

\textbf{Contributions:} In this paper we introduce a new bound for the expectation of the softplus function. Unlike other bounds, our proposal does not rely on extending the variational family, or introducing additional parameters to ensure the bound is tight. In fact, our bound is exact in the limit and can be truncated to any order to ensure a desired level of accuracy.

Subsequently we apply this new bound to variational logistic regression and (sparse) logistic Gaussian Process classification, referring to the resulting methods as Variational Inference with Probabilistic Error Reduction (VI-PER). Through extensive simulations we demonstrate that VI-PER leads to more accurate posterior approximations and improves on the well known issue of variance underestimation within the variational posterior, which can be of critical importance in real world applications as demonstrated in \Cref{sec:application} \citep{Blei2017, Durante2019}.

\newcommand{\propmethod}{VI--PER }
\newcommand{\propmethodGP}{VI--PER-GP }
\newcommand{\mcmethod}{VI--MC }
\newcommand{\pgmethod}{VI--PG }

\section{Proposal} \label{sec:method}

In this section we propose a new bound for the expectation of the softplus function, $\log(1 + \exp(X))$ where $X \sim N(\vartheta, \tau^2)$, and subsequently show how this bound can be used to compute a tight approximation to the ELBO in variational logistic regression and GP classification.

\subsection{A New Bound}

At its core variational logistic regression relies on the computation of a one dimensional integral, namely the expectation of the log-likelihood function,
$
    \E_{X} [
        y X
        - \log(1 + \exp(X))
    ]
$
for some uni-dimensional random variable $X$. However, this expectation is intractable as the softplus function $\log(1 + \exp(X))$ does not have a closed form integral. To this end, we propose a new bound for this expectation, which is summarized in the following theorem, a proof of which is given in Section \ref{appendix:proof_bound} of the Appendix.
\begin{theorem} \label{thrm:nb}
Let $X \sim N(\vartheta, \tau^2)$ then for any $l \geq 1$,
$\E_{X} [ \log(1 + \exp(X)) ] \leq \eta_l(\vartheta, \tau)$
where
\begin{equation} \label{eq:new_tight}
\begin{aligned}
    \eta_l(\vartheta, \tau) &
	= 
	    \frac{\tau}{\sqrt{2\pi}} e^{- \frac{\vartheta^2}{2 \tau^2}}
	    + \vartheta \Phi \left(\frac{\vartheta}{\tau} \right) 
	\\
	+ &\
	    \sum_{k=1}^{2l-1} 
		\frac{(-1)^{k-1}}{k} \Bigg[
		    e^{k \vartheta + k^2 \tau^2 / 2 }  \Phi \left( -\frac{\vartheta}{\tau} - k \tau \right) 
        \\
		  + & e^{-k \vartheta + k^2 \tau^2 /2 } \Phi \left( \frac{\vartheta}{\tau} -  k \tau \right)
		\Bigg] 
\end{aligned}
\end{equation}
and $\Phi(\cdot)$ is the cumulative distribution function of the standard normal distribution. 
\end{theorem} %

Notably, unlike the bound introduced by \citeauthor{Jakkola97} \yrcite{Jakkola97} (or the PG formulation), our bound does not rely on additional variational parameters, meaning no further optimization is necessary to guarantee tightness of the bound. In fact, irrespective of this, the proposed bound is at least as tight as that of \citeauthor{Jakkola97} (\citeyear{Jakkola97}) as seen in \Cref{fig:error_of_bound} (a) and (b), which is particularly evident when $\vartheta$ and $\tau$ are large (further corroborated in \Cref{appendix:error_of_bounds}). In turn, this means that the proposed bound is able to achieve a better approximation to the true expectation, which leads to more accurate posterior approximations as shown in Section \ref{sec:simulations}.

Furthermore, although $\eqref{eq:new_tight}$ is presented as a bound of the expectation, it is in fact exact in the limit when $l \to \infty$. This follows as a consequence of the following Lemma, a proof of which is given in Section \ref{appendix:limit} of the Appendix.
\begin{lemma} \label{lemma:asymptotic}
Let $a_k$ be the absolute value of the $k$-th term of the sum in Theorem \ref{thrm:nb}, then for $k \to \infty$ we have
\begin{equation} \label{eq:a_k}
\begin{aligned}
    a_k = \frac{1}{k}  \Bigg[
        & e^{k \vartheta}{}^{+\frac{k^2  \tau^2}{2}} \Phi \left(\! - \frac{\vartheta}{\tau}\! - \! k \tau \right) 
    \\
      & + e^{-k \vartheta + \frac{k^2 \tau^2}{2}} \Phi \left( \frac{\vartheta}{\tau}\! -\!  k \tau \right)
    \Bigg] 
    \sim \frac{1}{k^2} \underset{k \to \infty}{\to} 0
\end{aligned}
\end{equation}
\end{lemma}
As a result, \Cref{thrm:nb} converges to the true expectation in the limit, as summarized below.
\begin{corollary} \label{cor:bound}
Let 
\begin{equation}
S_{K} = \frac{\tau}{\sqrt{2\pi}} e^{- \frac{\vartheta^2}{2 \tau^2}}+ \vartheta \Phi \left(\frac{\vartheta}{\tau} \right) + \sum_{k=1}^K (-1)^{(k-1)} a_k.
\end{equation}
where $a_k$ is defined \eqref{eq:a_k}, then
\begin{equation}
    \lim_{K \to \infty} S_{2K} = \E_{X} \log(1 + \exp(X))
\end{equation}
\end{corollary}

In practice however, the sum is truncated at some $l \geq 1$, which can be chosen such that the relative error is below a given threshold. \Cref{fig:error_of_bound} (c) shows that a relative error below $1\%$ can be achieved when $l = 12$, which occurs about the origin when the variance $\tau^2$ is small. Further details on the choice of $l$ are given in Section \ref{appendix:impact_of_l} of the Appendix.


%

\subsection{Applications to Classification}

Two applications of \Cref{thrm:nb}, namely variational logistic regression and Gaussian process classification are presented next.  In both cases we show that the proposed bound can be used to compute a tight approximation to the ELBO without the need for additional parameters, costly Monte Carlo or quadrature methods.


\subsubsection{Variational Logistic Regression}

In the context of variational logistic regression $f(x) = x^\top \beta$ and $\beta \sim N_p(m, S)$ a priori where $m \in \R^p$ and $S \in \mathbb{S}_+^{p}$ are the prior mean and covariance respectively. Hence, inference involves approximating the posterior of $\beta$ with a distribution from the variational family $\Q' = N_d(\mu, \Sigma)$ with $d = p$, i.e. a single co-ordinate in the variational family is associated with a co-ordinate from the coefficient vector. Under this formulation the ELBO is given by,
\begin{equation} \label{eq:elbo_logistic}
\begin{aligned}
    \E_{q(\beta)} & \Bigg[  
        \sum_{i=1}^n 
        y_i x_i^\top  \beta 
        - \log(1 + \exp(x_i^\top \beta)) 
    \Bigg] \\ 
    -&\ \KL(q(\beta) || p(\beta))
\end{aligned}
\end{equation}
where
\begin{equation}
\begin{aligned}
    \KL(& q(\beta) \| p(\beta)) = \frac{1}{2} \big( \log \frac{|S|}{|\Sigma|} - p + \text{tr} (S^{-1}\Sigma) \\
     &\ + (\mu - m)^\top S^{-1} (\mu - m) \big).
\end{aligned}
\end{equation}

Using the fact that $ x_i^\top \beta \sim N(x_i^\top \mu, x_i^\top \Sigma x_i )$ the expectation of the softplus function in \eqref{eq:elbo_logistic} can be bounded by applying Theorem \ref{thrm:nb} with $\vartheta_i = x_i^\top \mu$ and $\tau_i^2 = x_i^\top \Sigma x_i$. Thus, giving a tractable lower bound to the ELBO of the form,
\begin{equation} \label{eq:elbo_logistic_bound}
\begin{aligned}
    \text{E}&\text{LBO}(q(\beta)) \geq 
    \F_l (\mu, \Sigma) 
    :=\\ &\  \sum_{i=1}^n \left( y_i x_i^\top \mu - \eta_l(\vartheta_i , \tau_i) \right)
    - \KL(q(\beta) || p(\beta)).
\end{aligned}
\end{equation}
In turn \eqref{eq:elbo_logistic_bound} can be maximized in place of the ELBO
with respect to the variational parameters $\mu$ and $\Sigma$ to give a surrogate variational posterior. 
This can be done in a number of ways e.g. via co-ordinate ascent variational inference or stochastic variational inference \cite{Blei2017,Zhang2019a}. Here we turn to gradient descent for simplicity by computing the gradient of $\F_l(\mu, \Sigma)$ with respect to $\mu$ and $\Sigma$ and updating the parameters accordingly. Notably, although \eqref{eq:elbo_logistic_bound} is a lower bound on the ELBO the use of surrogate lower bounds is a common technique in VI \cite{Komodromos2021, Depraetere2017a}.


\subsubsection{Gaussian Process Classification}

In the context of Logistic Gaussian Process classification 
a GP prior is placed on $f$, formally $f \sim \GP(m(\cdot), k(\cdot, \cdot) )$ where $m(\cdot)$ is the mean function and $k(\cdot, \cdot)$ is the kernel.
Inference now
involves computing the posterior distribution $\pi(f | \D)$, however, due to the lack of conjugacy and the fact that the computational complexity is $O(n^3)$, sparse variational inference 
is used to approximate the posterior
\citep{Titsias2009, Hensman2015, Wenzel2017}. 

In this vein we follow \citeauthor{Hensman2015} (\citeyear{Hensman2015}) and let the variational family be an $M$ dimensional Gaussian distribution, where $M$ are the number of inducing points (i.e. points used to perform the sparse approximation). Under this formulation the variational posterior is given by $q(u) = N_M(\mu, \Sigma)$ where $u$ are the inducing points and $\mu \in \R^M$ and $\Sigma \in \mathbb{S}_+^{M}$.

Using the fact that the random variables $u$ are points on the function in exactly the same way as $f$ are, the joint distribution can be written as
\begin{equation}
     p(f, u) = N \left(
         \begin{bmatrix}
             f \\ u
         \end{bmatrix}
         \middle|
         \begin{bmatrix}
             m(x) \\ m(z)
         \end{bmatrix}
         ,
         \begin{bmatrix}
             K_{nn} & K_{nm} \\
             K_{nm}^\top & K_{mm}
         \end{bmatrix}
     \right)
\end{equation}
where 
$K_{nn} = k(x, x)$, $K_{nm} = k(x, z)$, and $K_{mm} = k(z, z)$, where $z$ are the inducing point locations. In turn the ELBO with respect to $q(u)$ can be bounded by,
\begin{align}
    \text{ELBO}(&q(u)) =
    \E_{q(u)} \left[
        \log p(y | u)
    \right] 
    - \KL(q(u) || p(u))
    \nonumber \\
    \geq &\
    E_{q(u)}  
    \left[
        \E_{p(f | u)} \left[ \log (p(y | f)) \right]
    \right] 
    - \KL(q(u) || p(u)) 
    \nonumber \\
    = &\ 
    \E_{q(f)} \left[  \log p(y | f) \right] 
    - \KL(q(u) || p(u))
    \label{eq:gp_optim}
\end{align}
where $q(f) = N(A \mu, K_{nn} + A(\Sigma - K_{mm})A^\top)$ with $A = K_{nm} K_{mm}^{-1}$, and the inequality follows from the application of Jensen's inequality wherein,
$
     \log(p(y | u)) 
     = \log \left( \E_{p(f|u)} \left[ p(y | f) \right] \right) 
     \geq \E_{p(f|u)} \left[ \log(p(y | f)) \right]
$.

Given that the expectation of $\log(p(y | f))$ in \eqref{eq:gp_optim} is,
$$
      \E_{q(f)} \left[
         \sum_{i=1}^n 
         y_i f(x_i) 
         - \log(1 + \exp(f(x_i))) 
     \right],
$$
Theorem \ref{thrm:nb} can be applied to give a further lower bound on the ELBO,
\begin{equation} \label{eq:gp_bound}
\begin{aligned}
    \text{E}&\text{LBO}(q(u)) \geq 
    \F_l(\mu, \Sigma) := \\ &\ \sum_{i=1}^n \left( y_i m(x_i) - \eta_l(\vartheta_i, \tau_i) \right)
    - \KL(q(u) || p(u))
\end{aligned}
\end{equation}
where $\vartheta_i= (A \mu)_i$ and $\tau_i^2 = (K_{nn} + A(\Sigma - K_{mm})A^\top)_{ii}$ for $i=1,\dots,n$.
As before $\mathcal{F}_l(\mu, \Sigma)$ in \eqref{eq:gp_bound} is optimized using gradient descent to give the variational posterior. Notably, this can be done in conjunction with the inducing point locations and kernel hyperparameters if necessary (as is done in our implementation.)


\subsection{Computational Complexity}

The computational complexity is summarized in Table \ref{tab:complexity}. Notably, the proposed bound has computational complexity that depends on $l$, whereas the PG formulation (the probabilistic equivalent to \eqref{eq:jj_bound}) has a fixed complexity. However, the PG formulation uses $n$ additional parameters, as each data point has an associated variational parameter, which must be optimized as well. Whereas the proposed bound does not require any additional parameters, and so the number of parameters is fixed at $p$. This means a trade-off can be made between memory and computation time irregardless of methodological differences.

\vspace{-1em}
\bgroup
\def\arraystretch{1.05}%
\begin{table}[htp]
    \centering
    \caption{\textit{Computational and space complexity.} Complexity is given for a single observation.}
    \label{tab:complexity}
    \vspace{0.5em}
    \makebox[\columnwidth][c]{
    \resizebox{0.73\columnwidth}{!}{ %
    \begin{tabular}{ l  c c }
        \toprule
        \multicolumn{1}{c}{Method} & Parameters & Complexity \\
        \midrule
        Polya-Gamma &
        $p+n$ & $O(1)$ \\
        Our bound &
        $p$ & $O(2l - 1)$ \\
        \bottomrule
    \end{tabular}%
}}
    \vspace{-0.5em}
\end{table}
\egroup

\bgroup
\def\arraystretch{1.4}%
\begin{table*}[t]
\centering
\caption{\textit{Logistic regression results.} Median (2.5\%, 97.5\% quantiles) of the ELBO, KL$_\text{MC}$, MSE, coverage, CI width and AUC for the different methods. Here KL$_\text{MC}$ is the KL divergence between the posterior of $\beta$ computed via \mcmethod and the posterior computed via the respective method. Bold indicates the best performing variational method excluding \mcmethod which is considered the ground truth.}
\label{tab:logistic-reg-results}
\vspace{0.5em}
\makebox[\textwidth][c]{
\resizebox{1.0\textwidth}{!}{ %
\begin{tabular}{ c  c c  c c c c c c l}
    \toprule
    n / p & VF & Method & ELBO & KL$_\text{MC}$ & MSE & Coverage & CI Width & AUC & \multicolumn{1}{c}{Runtime} \\
    \midrule
    \multirow{7}{*}{1,000 / 25}
    & \multirow{3}{*}{Q}
    & \propmethod  & 
    \textbf{-264 (-340, -230)} & \textbf{0.00588 (0.0032, 0.01)} & \textbf{0.493 (0.19, 1.5)} & \textbf{0.906 (0.66, 0.99)} & \textbf{2.44 (2.1, 2.7)} & 0.977 (0.96, 0.98) & 12s (9.9s, 21s) \\
    && \mcmethod & 
-264 (-340, -230) & - & 0.494 (0.2, 1.5) & 0.904 (0.65, 0.99) & 2.42 (2.1, 2.7) & 0.977 (0.96, 0.98) & 2m 24s (1m 56s, 2m 57s) \\
    && \pgmethod & 
    -332 (-440, -280) & 5.36 (3.6, 6.9) & 0.557 (0.21, 1.7) & 0.724 (0.47, 0.92) & 1.67 (1.5, 1.8) & 0.977 (0.96, 0.98) & \textbf{0.12s (0.054s, 0.37s)} \\
    & \multirow{3}{*}{Q'}
    & \propmethod & 
    \textbf{-256 (-330, -220)} & \textbf{0.548 (0.42, 0.76)} & \textbf{0.493 (0.2, 1.5)} & \textbf{0.945 (0.72, 1)} & \textbf{2.65 (2.2, 3.1)} & 0.977 (0.96, 0.98) & 9.7s (3.8s, 28s) \\
    && \mcmethod & 
    -257 (-330, -220) & - & 0.491 (0.2, 1.5) & 0.949 (0.74, 1) & 2.66 (2.2, 3.1) & 0.977 (0.96, 0.98) & 2m 15s (1m 52s, 2m 49s) \\
    && \pgmethod & 
    -277 (-350, -240) & 8.15 (4.9, 12) & 0.554 (0.21, 1.7) & 0.723 (0.46, 0.93) & 1.66 (1.5, 1.8) & 0.977 (0.96, 0.98) & \textbf{0.57s (0.25s, 1.2s)} \\
    & \multicolumn{2}{c}{MCMC} & 
- & - & 0.492 (0.2, 1.5) & 0.948 (0.74, 1) & 2.66 (2.2, 3.1) & 0.977 (0.96, 0.98) & 10m 46s (6m 49s, 14m 57s) 
    \\ 
    \midrule
    \multirow{7}{*}{10,000 / 25}
    & \multirow{3}{*}{Q}
    & \propmethod & 
    \textbf{-2160 (-2900, -1700)} & \textbf{0.0341 (0.0066, 0.13)} & \textbf{0.0476 (0.025, 0.18)} & \textbf{0.918 (0.65, 0.98)} & \textbf{0.78 (0.67, 0.9)} & 0.974 (0.95, 0.98) & 53s (34s, 1m 35s) \\
    && \mcmethod & 
-2160 (-2900, -1700) & - & 0.0467 (0.026, 0.19) & 0.919 (0.65, 0.98) & 0.783 (0.67, 0.91) & 0.974 (0.95, 0.98) & 14m 15s (12m 49s, 15m 47s) \\
    && \pgmethod & 
    -3120 (-4100, -2400) & 4.89 (3.1, 7.6) & 0.0484 (0.026, 0.2) & 0.761 (0.46, 0.89) & 0.535 (0.49, 0.58) & 0.974 (0.95, 0.98) & \textbf{0.87s (0.48s, 1.6s)} \\
    & \multirow{3}{*}{Q'}
    & \propmethod & 
    \textbf{-2150 (-2900, -1700)} & \textbf{1.72 (1, 3.9)} & \textbf{0.0468 (0.026, 0.18)} & \textbf{0.96 (0.78, 0.99)} & \textbf{0.904 (0.73, 1.1)} & 0.974 (0.95, 0.98) & 1m 4.1s (24s, 1m 50s) \\
    && \mcmethod & 
-2160 (-2900, -1700) & - & 0.0475 (0.025, 0.18) & 0.971 (0.84, 1) & 0.958 (0.77, 1.2) & 0.974 (0.95, 0.98) & 13m 49s (10m 1.7s, 15m 33s) \\
    && \pgmethod & 
    -2170 (-2900, -1700) & 12.6 (7.5, 21) & 0.0483 (0.026, 0.2) & 0.764 (0.46, 0.9) & 0.539 (0.49, 0.58) & 0.974 (0.95, 0.98) & \textbf{3.9s (2.3s, 7.5s)} \\
    & \multicolumn{2}{c}{MCMC} & 
- & - & 0.0469 (0.026, 0.19) & 0.959 (0.77, 0.99) & 0.89 (0.71, 1.1) & 0.974 (0.95, 0.98) & 18m 3.1s (12m 41s, 20m 44s) 
    \\ 
    \bottomrule
\end{tabular}%
}}
\end{table*}
\egroup

\subsection{Implementation Details}

To ensure stable optimization a re-parameterization of the variational parameters is used. In the context of logistic regression, 
we let $\Sigma = L L^\top$ where $L$ is a lower triangular matrix and optimize over the elements of $L$. In terms of $\mu$ no re-parametrization is made. For Logistic Gaussian process classification the parameterization 
$\theta = \Sigma^{-1} \mu$ and $\Theta = -\frac{1}{2} \Sigma^{-1}$
is made, and optimization is performed over $\theta$ and $\Theta$. The variational parameters are then recovered via $\mu = \Sigma \theta$ and $\Sigma = -2 \Theta^{-1}$. Beyond ensuring stable optimization, this parameterization gives rise to natural gradients, known to lead to faster convergence \citep{Martens2020}.

Regarding the initialization of $\mu$ and $\Sigma$, the mean vector $\mu$ is sampled from a Gaussian distribution with zero mean and identity covariance matrix, and $\Sigma = 0.35 I_p$. To assess convergence the relative change in the ELBO is monitored, given by $ \Delta \text{ELBO}_t = |\text{ELBO}_t - \text{ELBO}_{t-1}|/|\text{ELBO}_{t-1}|$. Once this quantity is below a given threshold, the gradient descent algorithm is stopped. In practice we find that a threshold between $10^{-6}$ and $10^{-8}$ is sufficient.

Finally, we note our implementation is based on PyTorch \citep{Paszke2019} and uses Gpytorch \citep{Gardner2018} to perform Gaussian Process VI. The implementation 
is
freely available 
at \url{https://github.com/mkomod/vi-per}.

\section{Numerical Experiments} \label{sec:simulations}

In this section a numerical evaluation of our method taking $l=12$ is performed. Referring to our method as Variational Inference with Probabilistic Error Reduction (\propmethod\!\!), we compare against the Polya-Gamma formulation (\pgmethod\!\!) [which is a probabilistic interpretation of the bound introduced by \citeauthor{Jakkola97} (\citeyear{Jakkola97})] and the ELBO computed via Monte-Carlo (\mcmethod\!\!) using 1,000 samples. Throughout we consider the variational posterior computed with Monte Carlo as the ground truth and use this as a reference to evaluate the performance of the other methods.

Furthermore in the case of variational logistic regression a further comparison to the posterior distribution obtained via MCMC is made. Notably, Hamiltonian Monte Carlo is used to sample from the posterior which is implemented using Hamiltorch \citep{cobb2020scaling}. For our sampler we use 30,000 iterations and a burn-in of 25,000 iterations. The step size is set to 0.01 and the number of leapfrog steps is set to 25. For the Gaussian process classification we do not compare to MCMC due to the high computational cost of sampling from the posterior \cite{RasmussenW06}.

To evaluate the performance of the methods we report:
\vspace{-0.5em}
\begin{enumerate}[label=(\roman*)]
    \itemsep0em
    \item The ELBO estimated via Monte-Carlo (using 10,000 samples) to ensure consistency across methods.
    \item The KL divergence between the posterior obtained via \mcmethod and the respective method, denoted by $\text{KL}_{\text{MC}}$.
    \item The mean squared error (MSE) between the posterior mean of $f(x_i)$ and the value of the true model $f_0(x_i)$ for $i=1,\dots,n$.
    \item The coverage of the 95\% credible interval (CI), which is the proportion of times $f_0(x_i)$ is contained in the marginal credible interval of $f(x_i)$ for $i=1,\dots,n$.
    \item The width of the 95\% CI of $f(x_i)$.
    \item The area under the curve (AUC) of the receiver operating characteristic (ROC) curve, which is a measure of the predictive performance of the model.
\end{enumerate}
\vspace{-0.5em}
Notably, we report the median and 2.5\% and 97.5\% quantiles of these metrics across 100 runs. Finally, details of the computational environment are given in Section \ref{appendix:computational_environment} of the Appendix.

\subsection{Logistic Regression Simulation Study} \label{sec:logistic-regression}

Our first simulation study evaluates the performance of VI-PER in the context of variational logistic regression. To this end, we consider datasets with $n=1,\!000$ and $n=10,\!000$ observations, and $p=25$ predictors. Additional results of varying values of $n$, $p$, and predictor sampling schemes are presented in Section \ref{appendix:logistic_regression} of the Appendix.

Here data is simulated for $i=1,\dots,n$ observations each having a response $y_i \in \{0, 1\}$ and $p$ continuous predictors $x_i \in R^p$. The response is sampled independently from a Bernoulli distribution with parameter $p_i = 1/(1 + \exp(-x_i^\top \beta_0))$ where the true coefficient vector $\beta_0 = (\beta_{0, 1}, \dots, \beta_{0, p})^\top \in \R^p$ which elements $\beta_{0, j} \overset{\text{iid}}{\sim} U([-2.0, 0.2] \cup [0.2, 2.0])$ for $j=1,\dots,p$. Finally, the predictors 
$
    x_i \overset{\text{iid}}{\sim} N(0_p, W^{-1})$ where $W \sim \text{Wishart}(p + 3, I_p)
$, which ensures that the predictors are correlated.

\begin{figure*}[ht]
    \centering
\makebox[\textwidth][c]{
\resizebox{1.0\textwidth}{!}{ %
    \includegraphics[width=1.00\textwidth]{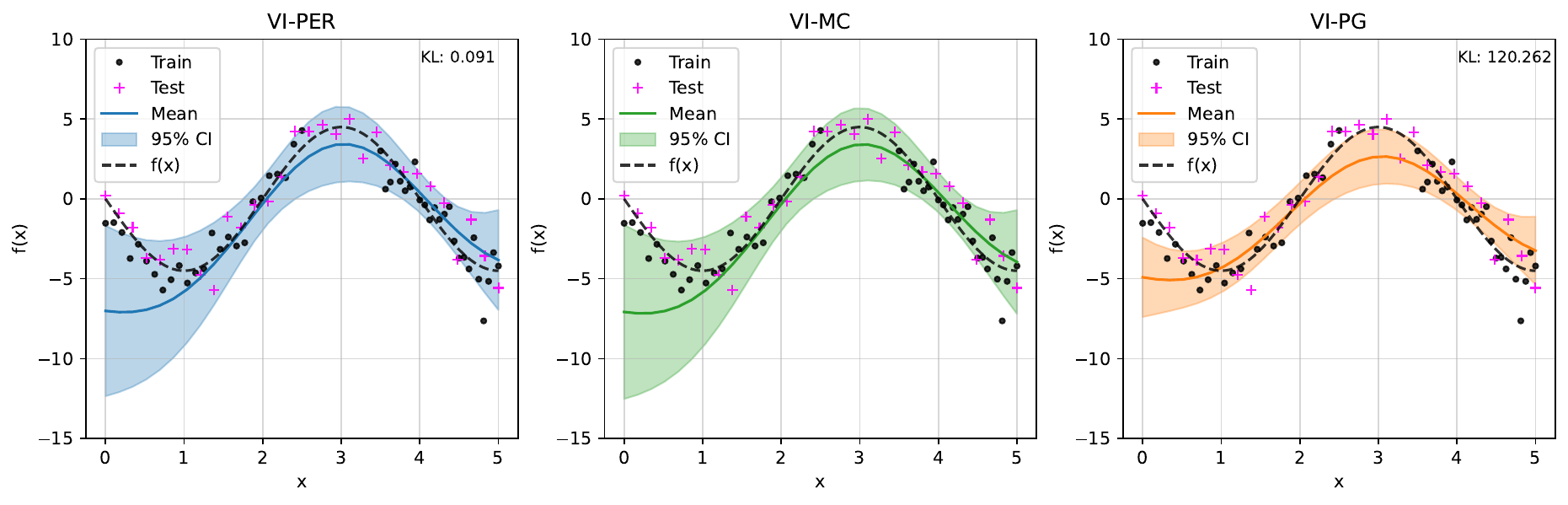} %
}}
    \vspace{-2em}
    \caption{\textit{GP classification: illustrative example}. Presented is the mean (solid line) and 95\% credible interval (shaded region) of the posterior distribution for the different methods. The true function is shown in dashed line (\dashedline{black}), the training data are given by the black points (\pointFilled{black}) and the test data by the magenta crosses ({\color{magenta} +}). In the top right corner the KL divergence between the variational posterior computed using Monte Carlo and the variational posterior computed using the respective method is presented.}
    \label{fig:toy-example}
    \vspace{-0.5em}
\end{figure*}

Highlighted in Table \ref{tab:logistic-reg-results} are the results for the different methods, these show that \propmethod is able to achieve similar performance to \mcmethod (considered the ground truth amongst the variational methods), while being significantly faster to compute. Furthermore, \propmethod is able to achieve similar predictive performance as with \pgmethod, however our method shows significant improvements across several metrics. In particular, \propmethod obtains a lower MSE, higher coverage and larger CI width, meaning that \propmethod is able to achieve a better fit to the data and a more accurate representation of the posterior uncertainty. This is made particularly evident as the $\text{KL}_{\text{MC}}$ for \propmethod is significantly lower than that of \pgmethod\!\!.

Finally, although we do not consider a divergence between the variational posterior and the true posterior (as we are unable to compute $\KL(\Pi(\beta | \D) \| q(\beta))$ due to the unknown normalizing constant), we note that the MSE, coverage and CI width are comparable to those of MCMC (considered the gold standard in Bayesian inference). This indicates that the variational posterior computed via \propmethod is an excellent approximation to the true posterior.

\subsection{Gaussian Process Classification: Illustrative Example} \label{sec:gp-logistic}

Our second simulation study is illustrative and used to demonstrate the
performance of \propmethod in the context of GP classification. Further evaluations are presented in Section \ref{sec:application} where \propmethod is applied to real data sets. In all our applications we consider a GP model with $M=50$ inducing points, linear mean function and ARD kernel with lengthscales initialized at $0.5$. 

In this setting, data is generated for $i=1,\dots,50$ samples, with $y_i \sim \text{Bernoulli}(p_i)$ where $p_i = s(f(x_i) + \epsilon_i)$, $f(x_i) = -4.5 \sin(  \frac{\pi}{2} x_i )$ and $\epsilon_i \sim N(0, 1)$. Here the predictors ($x_i$s) are given by a grid of points spaced evenly over $[0, 5] \setminus [2.5, 3.5]$. A test dataset of size $n=50$ is generated in the same way, however the predictors are evenly spaced over $[0, 5]$, meaning that the test data contains points in the interval $[2.5, 3.5]$ which are not present in the training data.

Figure \ref{fig:toy-example} illustrates a single realization of the synthetic data. The figure highlights, that \propmethod obtains a similar fit to the data as with \mcmethod (which is considered the ground truth amongst the variational methods). Furthermore, \Cref{fig:toy-example} showcases that the variational posterior variance is underestimated by \pgmethod\!\!, meaning that the CI width is too small. As a result the method fails to capture most of the points in the interval $[2.5, 3.5]$.

This statement is further supported by the results presented in Table \ref{tab:logistic-gp-results} where the simulation is repeated 100 times.
Notably, \propmethod shows improvements in the estimation of $f$, in particular the $\text{KL}_{\text{MC}}$ and MSE is lower, whilst the coverage is higher. These metrics suggest that \propmethod performs similarly with \mcmethod which is considered the baseline amongst the variational methods. Beyond this the runtime of our method is slightly lower, which is attributed to the fact that fewer iterations are needed to achieve convergence which on average is 453, 1290 and 926 iterations for \propmethod, \mcmethod and \pgmethod respectively. Furthermore, the proposed bound is able to achieve similar predictive performance as with the \pgmethod and \mcmethod formulation in terms of the AUC.

\begin{figure*}[ht]
    \centering
\makebox[\textwidth][c]{
\resizebox{1.0\textwidth}{!}{ %
    \includegraphics[width=1.00\textwidth]{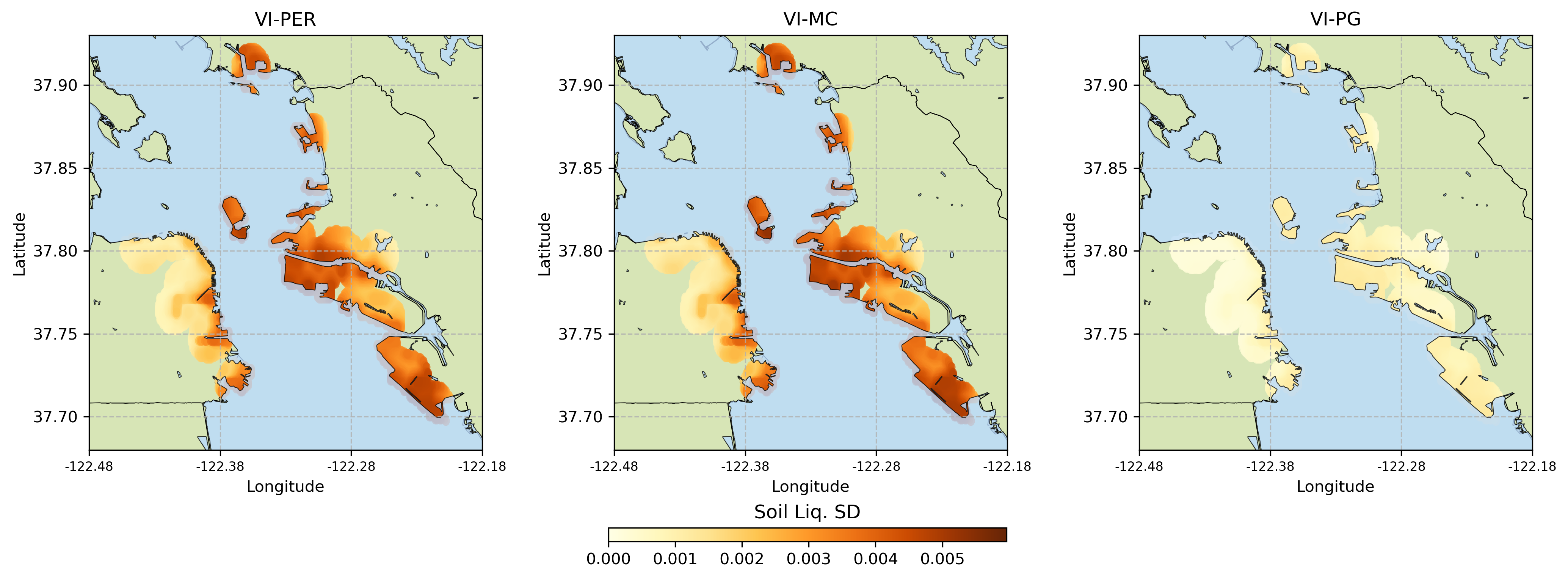} %
}}
    \vspace{-2em}
    \caption{\textit{Application to Soil liquefaction}. Standard deviation of soil liquefaction probability evaluated for the Loma Prieta earthquake for \propmethod, \mcmethod and \pgmethod under the variational family $\Q$.}
    \label{fig:lomaprieta}
\end{figure*}

\vspace{-1.0em}
\bgroup
\def\arraystretch{1.4}%
\begin{table}[h]
\caption{\textit{GP Classification: illustrative example results.} Median (2.5\%, 97.5\% quantiles) of the ELBO, KL$_\text{MC}$, MSE, coverage, CI width and AUC for the different methods. Here \textit{grid} refers to these quantities computed along a grid of values in $[0, 5]$.}
\label{tab:logistic-gp-results}
\vspace{0.5em}
    \centering
\makebox[\columnwidth][c]{
\resizebox{1.0\columnwidth}{!}{ %
\begin{tabular}{ l  l  l  l }
    \toprule
    \multicolumn{1}{c}{} & \multicolumn{1}{c}{\propmethod} & \multicolumn{1}{c}{\mcmethod} & \multicolumn{1}{c}{\pgmethod} \\
    \midrule
    ELBO & 
    \textbf{-24.1 (-31, -16)} & -24.5 (-31, -16) & -24.5 (-29, -17) \\
    KL$_{\text{MC}}$ (grid) &
    \textbf{2.78 (0.071, 130)} & \multicolumn{1}{c}{-} & 37.5 (9.7, 270) \\
    MSE (grid) & 
    \textbf{1.95 (0.42, 8.1)} & 1.93 (0.49, 7.8) & 2.25 (0.65, 7.8) \\
    CI width (grid) &
    \textbf{4.31 (2.4, 6.9)} & 4.22 (2.6, 6.6) & 3.13 (2.1, 3.9) \\
    Coverage (grid) &
    \textbf{0.89 (0.2, 1)} & 0.88 (0.29, 1) & 0.68 (0.21, 1) \\
    AUC (test) & 
    \textbf{0.928 (0.72, 0.99)} & 0.921 (0.76, 0.99) & 0.923 (0.75, 0.99) \\
    Runtime & 
    \textbf{10s (3.7s, 25s)} & 39s (4s, 41s) & 18s (4.5s, 34s) \\
    \bottomrule
\end{tabular} %
}}
\end{table}
\egroup

\section{Application to Real World Data} \label{sec:application}

Finally, we present two applications of \propmethod to real world data. The first is to the problem of soil liquefaction, which illustrates the necessity of scalable uncertainty quantification in real world settings. The second application is to a number of publicly available datasets and is used to further evaluate the performance of GP classification with \propmethod.

\subsection{Application to Soil Liquefaction Data}

The first application is to the problem of soil liquefaction, a phenomenon that occurs when saturated soil loses strength and stiffness due to an earthquake. Soil liquefaction is a
secondary hazard of earthquakes and can cause ground failure and severe structural damage.
Thus, understanding the probability of soil liquefaction is vital for risk assessment, mitigation and emergency response planning \cite{Zhan2023}.

To model soil liquefaction we use data from a study by \citeauthor{Zhan2023} (\citeyear{Zhan2023}), which was accessed with permission of the author and will be publicly available in the near future. The dataset consists of data from 25 earthquakes that took place between 1949 -- 2015. In total there are 1,809,300 observations collected at different locations for each earthquake, which consist of 33 features and a binary response indicating whether or not soil liquefaction occurred at a given location. We follow \citeauthor{Zhan2023} (\citeyear{Zhan2023}) and construct a model consisting of five features, namely:
\vspace{-.5em}
\begin{enumerate}[label=(\roman*)]
    \itemsep0em
    \item Peak ground velocity, which is a measure of the maximum velocity of the ground during an earthquake.
    \item Shear wave velocity within the top 30 m of the soil column. 
    \item Mean annual precipitation.
    \item The distance to the nearest water body.
    \item The ground water table depth.
\end{enumerate}

Following \citeauthor{Zhan2023} \yrcite{Zhan2023} models are trained using 24 of the earthquakes and tested on the remaining earthquake which took place is Loma Prieta in 1989. Notably the training set consists of 1,719,400 samples and the test set consists of 89,900 samples. The results are presented in Table \ref{tab:soil_liquefaction_results} and show that \propmethod is able to achieve similar predictive performance to the \pgmethod and \mcmethod in terms of the AUC. However \propmethod obtains a higher ELBO suggesting a better fit to the data. Furthermore, \propmethod obtains wider CI widths inline with \mcmethod\!\!, suggesting \pgmethod is underestimating the posterior uncertainty.

This is made particularly evident in Figure \ref{fig:lomaprieta}, which shows the standard deviation of probability of soil liquefaction for the Loma Prieta earthquake. The figure highlights that \propmethod propagates the uncertainty in the data inline with \mcmethod\!\!, whereas \pgmethod underestimates this quantity. Overall, these results suggest that
\propmethod can provide tangible benefits in real world settings where uncertainty quantification is of vital importance.

\bgroup
\def\arraystretch{1.4}%
\begin{table*}[htp]
\centering
\caption{\textit{GP classification: application to real data.} Median (2.5\%, 97.5\% quantiles) of the ELBO, KL$_\text{MC}$, CI width and AUC. Here KL$_\text{MC}$ is the KL divergence between the posterior of $f(x_i)$ computed via \mcmethod and the posterior computed via the respective method evaluated at the test/training data. Bold indicates the best performing method excluding \mcmethod which is considered the ground truth.}
\vspace{0.5em}
\label{tab:logistic-reg-app}
\makebox[\textwidth][c]{
\resizebox{1.0\textwidth}{!}{ %
    \begin{tabular}{ l c  p{1.5cm}  l l l l l l }
    \toprule
    Dataset & n / p & \multicolumn{1}{c}{Method} & \multicolumn{1}{c}{ELBO} & \multicolumn{1}{c}{KL$_{\text{MC}}$ (train)} & \multicolumn{1}{c}{KL$_{\text{MC}}$ (test)} & \multicolumn{1}{c}{CI width (test)} & AUC (test) & \multicolumn{1}{c}{Runtime}
    \\ 
    \midrule
\multirow{3}{*}{breast-cancer} & \multirow{3}{*}{683 / 10} &
\multicolumn{1}{c}{\propmethod} &
        \textbf{-41.83 (-45.06, -39.73)} & \textbf{102 (3.34, 826)} & \textbf{12.4 (0.339, 88.8)} & \textbf{5.92 (4.38, 7.08)} & 0.995 (0.992, 0.998) & 25s (13s, 53s) \\
&& \multicolumn{1}{c}{\mcmethod} &
-41.87 (-45.55, -39.27) & \multicolumn{1}{c}{-} & \multicolumn{1}{c}{-} & 6.16 (3.81, 7.37) & 0.995 (0.991, 0.997) & 2m 9.9s (29s, 2m 16s) \\
&& \multicolumn{1}{c}{\pgmethod} & 
-47.09 (-50.58, -46.97) & 217000 (67900, 1.84e+06) & 23000 (7350, 187000) & 0.44 (0.173, 0.849) & \textbf{0.997 (0.994, 0.998)} & \textbf{14s (3.7s, 44s)} \\
\midrule
\multirow{3}{*}{svmguide1} & \multirow{3}{*}{3089 / 4} &
\multicolumn{1}{c}{\propmethod} &
        \textbf{-267.1 (-295.8, -248.9)} & \textbf{1290 (50.7, 18500)} & \textbf{1580 (64.4, 20000)} & \textbf{5.11 (3.64, 5.67)} & 0.996 (0.996, 0.996) & \textbf{1m 37s (20s, 4m 8.6s)} \\
&& \multicolumn{1}{c}{\mcmethod} &
-266.5 (-306.5, -250) & \multicolumn{1}{c}{-} & \multicolumn{1}{c}{-} & 5.18 (3.65, 5.46) & 0.996 (0.996, 0.996) & 10m 21s (1m 42s, 10m 52s) \\
&& \multicolumn{1}{c}{\pgmethod} &
-285.8 (-330.4, -263.6) & 101000 (40400, 301000) & 111000 (49200, 307000) & 1.51 (1.46, 1.6) & 0.996 (0.996, 0.996) & 1m 53s (15s, 4m 7.8s) \\

\midrule
\multirow{3}{*}{australian} & \multirow{3}{*}{690 /  14 } &
\multicolumn{1}{c}{\propmethod} &
\textbf{-191.2 (-194.3, -186.9)} & \textbf{1020 (41.4, 97200)} & \textbf{170 (6.89, 23000)} & \textbf{1.56 (0.194, 2.55)} & \textbf{0.953 (0.945, 0.961)} & 25s (12s, 1m 6.9s) \\
&& \multicolumn{1}{c}{\mcmethod} &
-192.2 (-198.9, -185.9) & \multicolumn{1}{c}{-} & \multicolumn{1}{c}{-} & 1.49 (0.279, 2.61) & 0.951 (0.938, 0.959) & 1m 2.6s (17s, 2m 23s) \\
&& \multicolumn{1}{c}{\pgmethod} &
-193.9 (-195.1, -191.7) & 4020 (129, 74300) & 1090 (24.3, 29500) & 0.627 (0.194, 1.09) & 0.95 (0.947, 0.955) & \textbf{14s (3.9s, 55s)} \\

\midrule
\multirow{3}{*}{fourclass} & \multirow{3}{*}{862 / 2} &
\multicolumn{1}{c}{\propmethod} &
\textbf{-52.04 (-67.08, -47.92)} & \textbf{75 (4.78, 592)} & \textbf{7.81 (0.479, 64.3)} & \textbf{8.6 (6.04, 9.69)} & 1 (1, 1) & 57s (13s, 1m 24s) \\
&& \multicolumn{1}{c}{\mcmethod} &
-55.54 (-68.03, -53.4) & \multicolumn{1}{c}{-} & \multicolumn{1}{c}{-} & 7.74 (5.99, 8.15) & 1 (1, 1) & 2m 31s (59s, 2m 47s) \\
&& \multicolumn{1}{c}{\pgmethod} &
-71 (-84.94, -69.3) & 3850 (2870, 5620) & 384 (326, 604) & 2.95 (2.71, 3) & 1 (1, 1) & \textbf{51s (12s, 1m 14s)} \\
\midrule
\multirow{3}{*}{heart} & \multirow{3}{*}{270 / 13} &
\multicolumn{1}{c}{\propmethod} &
\textbf{-77.74 (-79.19, -75.72)} & \textbf{417 (21.6, 66800)} & \textbf{45.2 (2.67, 7510)} & \textbf{2.19 (0.202, 3.53)} & 0.894 (0.867, 0.922) & 12s (4.8s, 33s) \\
&& \multicolumn{1}{c}{\mcmethod} &
-77.51 (-79.98, -75.59) & \multicolumn{1}{c}{-} & \multicolumn{1}{c}{-} & 2.71 (0.219, 3.81) & 0.894 (0.867, 0.922) & 51s (9.5s, 1m 25s) \\
&& \multicolumn{1}{c}{\pgmethod} &
-78.93 (-79.44, -77.31) & 2200 (9.21, 11700) & 262 (1.01, 1420) & 0.713 (0.401, 1.43) & 0.894 (0.883, 0.9) & \textbf{5.1s (2.7s, 15s)} \\
\bottomrule
\end{tabular}%
}}
\end{table*}
\egroup

\vspace{-1em}
\bgroup
\def\arraystretch{1.4}%
\begin{table}[H]
    \centering
\caption{\textit{Soil Liquefaction results.} Evaluation of the ELBO, KL$_\text{MC}$, CI width and AUC for the different methods.}
\label{tab:soil_liquefaction_results}
\vspace{0.5em}
\makebox[\columnwidth][c]{
\resizebox{0.85\columnwidth}{!}{ %
\begin{tabular}{ c l  l c c c }
    \toprule
    $\quad$VF$\quad$ & Method & ELBO & KL$_\text{MC}$ & CI width & AUC (test) \\
    \midrule
    \multirow{3}{*}{$\Q$} &
    \propmethod &
\textbf{-835200} & \textbf{1.83} & \textbf{0.0335} & \textbf{0.858} \\
    & \mcmethod &
-835300 & - & 0.0335 & 0.857 \\
    & \pgmethod &
-921600 & 73.49 & 0.0065 & 0.857 \\
    \midrule
    \multirow{3}{*}{$\Q'$} &
    \propmethod &
-835200 & 9.11 & \textbf{0.0129} & \textbf{0.858} \\
    & \mcmethod &
-835200 & - & 0.0121 & 0.857 \\
    & \pgmethod &
-835200 & \textbf{4.82} & 0.0101 & 0.857 \\
    \bottomrule
\end{tabular}%
}}
\end{table}
\egroup

\subsection{Application to Publicly Available Datasets}


Here logistic Gaussian Process classification is applied to a number of publicly available datasets, all of which accessible through UCI or the LIBSVM package \citep{Chang2011}. The datasets summarized in Table \ref{tab:logistic-reg-app} include a number of binary classification problems with varying numbers of observations and predictors.

For each dataset we use the first 80\% of the data for training and the remaining 20\% for testing (when a testing set is not available). To evaluate the performance of the different methods the same metrics as in Section \ref{sec:simulations} are used, namely the ELBO, KL$_\text{MC}$, CI width and AUC. Noting, that the MSE and coverage are not reported as the true function is unknown. As before, the median and 2.5\% and 97.5\% quantiles of these metrics across 100 runs is reported.

The results presented in Table \ref{tab:logistic-reg-app} show that \propmethod is able to achieve similar predictive performance to \pgmethod in terms of the AUC. However \propmethod obtains a higher ELBO suggesting a better fit to the data. Furthermore, \propmethod obtains CI widths inline with \mcmethod indicating that \propmethod is able to capture the posterior uncertainty more accurately. As in earlier sections the KL divergence between \mcmethod and \propmethod is significantly lower than that of \mcmethod and \pgmethod\!\!, meaning that \propmethod is in closer agreement with \mcmethod\!\!, considered the ground truth amongst the methods.

\section{Discussion} \label{sec:discussion}

We have developed a novel bound for the expectation of the softplus function, and subsequently applied this to variational logistic regression and Gaussian process classification. Unlike other approaches, ours does not rely on extending the variational family, or introducing additional parameters to ensure the approximation is tight. 

Through extensive simulations we have demonstrated that our proposal leads to more accurate posterior approximations, improving on the well known issue of variance underestimation within the variational posterior \citep{Durante2019}. Furthermore, we have applied our method to a number of real world datasets, including a large dataset of soil liquefaction. An application which highlights the necessity of scalable uncertainty quantification, and demonstrates that our bound is able to achieve similar performance to the Polya-Gamma formulation in terms of the AUC, while significantly improving on the uncertainty quantification.

However, our method is not without its limitations. In particular, the proposed bound must be truncated, introducing error into the computation of the ELBO, and as a result the variational posterior. Furthermore, as with all variational methods, the variational family may not be flexible enough to approximate the true posterior, for example if there are multimodalities or heavy tails. As such, practitioners should take care when using our method, and ensure that the resulting posterior is sufficiently accurate for their application.

Finally, we note that there are several potential avenues of methodological application of our bound in many areas of machine learning, including: Bayesian Neural Network classification, logistic contextual bandits and Bayesian optimization with binary auxiliary information \citep{Z2019}, noting that the later two applications heavily rely on accurate posterior uncertainty quantification. Furthermore, various extensions can be made to the proposed method, including the use of more complex variational families such as mixtures of Gaussians.



\section*{Software and Data}

The code for the experiments in this paper is available at \url{https://github.com/mkomod/vi-per}

\section*{Acknowledgements}
The authors would like to thank the reviewers for their helpful comments and suggestions. This work was supported by the EPSRC Centre for Doctoral Training in Statistics and Machine Learning (EP/S023151/1), Imperial College London's Cancer Research UK centre and Imperial College London's Experimental Cancer Medicine centre. The authors would also like to thank the Imperial College Research Computing Service for providing computational resources and support that have contributed to the research results reported within this paper.

\section*{Impact Statement}
This paper presents work whose goal is to provide improved uncertainty quantification, enabling safer and more reliable decision making. The proposed method is applicable to a wide range of problems, both in applied fields and in machine learning. Ultimately, this work will enable the use of Bayesian methods in real world applications where uncertainty quantification is of critical importance, particularly when the computational cost of existing methods is prohibitive or there are time constraints on the decision making process e.g. medical diagnosis, robotics, natural disasters and autonomous vehicles.

However, we note our method is not without its limitations and provides approximate inference. As such practitioners should take care when using our method, and ensure that the resulting posterior is sufficiently accurate for their application.

\bibliography{references}

\begin{thebibliography}{27}
\providecommand{\natexlab}[1]{#1}
\providecommand{\url}[1]{\texttt{#1}}
\expandafter\ifx\csname urlstyle\endcsname\relax
  \providecommand{\doi}[1]{doi: #1}\else
  \providecommand{\doi}{doi: \begingroup \urlstyle{rm}\Url}\fi

\bibitem[Bishop(2007)]{bishop2007}
Bishop, C.~M.
\newblock \emph{Pattern Recognition and Machine Learning (Information Science and Statistics)}.
\newblock Springer, 2007.
\newblock ISBN 0387310738.

\bibitem[Blei et~al.(2017)Blei, Kucukelbir, and McAuliffe]{Blei2017}
Blei, D.~M., Kucukelbir, A., and McAuliffe, J.~D.
\newblock {Variational Inference: A Review for Statisticians}.
\newblock \emph{Journal of the American Statistical Association}, 112\penalty0 (518):\penalty0 859--877, 2017.
\newblock ISSN 1537274X.
\newblock \doi{10.1080/01621459.2017.1285773}.

\bibitem[Chang \& Lin(2011)Chang and Lin]{Chang2011}
Chang, C.~C. and Lin, C.~J.
\newblock {LIBSVM: A Library for support vector machines}.
\newblock \emph{ACM Transactions on Intelligent Systems and Technology}, 2\penalty0 (3):\penalty0 1--40, 2011.
\newblock ISSN 21576904.
\newblock \doi{10.1145/1961189.1961199}.

\bibitem[Chen et~al.(2021)Chen, Nie, and Li]{Xin2021}
Chen, X., Nie, Y., and Li, N.
\newblock Online residential demand response via contextual multi-armed bandits.
\newblock \emph{IEEE Control Systems Letters}, 5\penalty0 (2):\penalty0 433--438, 2021.
\newblock \doi{10.1109/LCSYS.2020.3003190}.

\bibitem[Cobb \& Jalaian(2021)Cobb and Jalaian]{cobb2020scaling}
Cobb, A.~D. and Jalaian, B.
\newblock Scaling hamiltonian monte carlo inference for bayesian neural networks with symmetric splitting.
\newblock \emph{Uncertainty in Artificial Intelligence}, 2021.

\bibitem[Depraetere \& Vandebroek(2017)Depraetere and Vandebroek]{Depraetere2017a}
Depraetere, N. and Vandebroek, M.
\newblock {A comparison of variational approximations for fast inference in mixed logit models}.
\newblock \emph{Computational Statistics}, 32\penalty0 (1):\penalty0 93--125, 2017.
\newblock ISSN 16139658.
\newblock \doi{10.1007/s00180-015-0638-y}.

\bibitem[Durante \& Rigon(2019)Durante and Rigon]{Durante2019}
Durante, D. and Rigon, T.
\newblock {Conditionally conjugate mean-field variational Bayes for logistic models}.
\newblock \emph{Statistical Science}, 34\penalty0 (3):\penalty0 472--485, 2019.
\newblock ISSN 21688745.
\newblock \doi{10.1214/19-STS712}.

\bibitem[Gardner et~al.(2018)Gardner, Pleiss, Bindel, Weinberger, and Wilson]{Gardner2018}
Gardner, J.~R., Pleiss, G., Bindel, D., Weinberger, K.~Q., and Wilson, A.~G.
\newblock {Gpytorch: Blackbox matrix-matrix Gaussian process inference with GPU acceleration}.
\newblock \emph{Advances in Neural Information Processing Systems}, 31\penalty0 (NeurIPS):\penalty0 7576--7586, 2018.
\newblock ISSN 10495258.

\bibitem[Gibbs \& MacKay(2000)Gibbs and MacKay]{Gibbs2000}
Gibbs, M.~N. and MacKay, D.~J.
\newblock {Variational Gaussian process classifiers}.
\newblock \emph{IEEE Transactions on Neural Networks}, 11\penalty0 (6):\penalty0 1458--1464, 2000.
\newblock ISSN 10459227.
\newblock \doi{10.1109/72.883477}.

\bibitem[Giordano et~al.(2018)Giordano, Broderick, and Jordan]{Giordano2018}
Giordano, R., Broderick, T., and Jordan, M.~I.
\newblock {Covariances, robustness, and variational Bayes}.
\newblock \emph{Journal of Machine Learning Research}, 19:\penalty0 1--49, 2018.
\newblock ISSN 15337928.

\bibitem[Hau{\ss}mann et~al.(2017)Hau{\ss}mann, Hamprecht, and Kandemir]{Haubmann2017}
Hau{\ss}mann, M., Hamprecht, F.~A., and Kandemir, M.
\newblock {Variational Bayesian multiple instance learning with Gaussian processes}.
\newblock In \emph{Proceedings - 30th IEEE Conference on Computer Vision and Pattern Recognition, CVPR 2017}, volume 2017-Janua, pp.\  810--819, 2017.
\newblock ISBN 9781538604571.
\newblock \doi{10.1109/CVPR.2017.93}.

\bibitem[Hensman et~al.(2015)Hensman, Matthews, and Ghahramani]{Hensman2015}
Hensman, J., Matthews, A.~G., and Ghahramani, Z.
\newblock {Scalable variational Gaussian process classification}.
\newblock \emph{Journal of Machine Learning Research}, 38:\penalty0 351--360, 2015.
\newblock ISSN 15337928.

\bibitem[Jaakkola \& Jordan(2000)Jaakkola and Jordan]{Jakkola97}
Jaakkola, T.~S. and Jordan, M.~I.
\newblock {Bayesian parameter estimation via variational methods}.
\newblock \emph{Statistics and Computing}, 10\penalty0 (1):\penalty0 25--37, 2000.
\newblock ISSN 09603174.
\newblock \doi{10.1023/A:1008932416310}.

\bibitem[Komodromos et~al.(2022)Komodromos, Aboagye, Evangelou, Filippi, and Ray]{Komodromos2021}
Komodromos, M., Aboagye, E.~O., Evangelou, M., Filippi, S., and Ray, K.
\newblock {Variational Bayes for high-dimensional proportional hazards models with applications within gene expression}.
\newblock \emph{Bioinformatics}, 38\penalty0 (16):\penalty0 3918--3926, aug 2022.
\newblock ISSN 1367-4803.
\newblock \doi{10.1093/bioinformatics/btac416}.
\newblock URL \url{http://arxiv.org/abs/2112.10270 https://academic.oup.com/bioinformatics/article/38/16/3918/6617825}.

\bibitem[Komodromos et~al.(2023)Komodromos, Evangelou, Filippi, and Ray]{komodromos2023}
Komodromos, M., Evangelou, M., Filippi, S., and Ray, K.
\newblock Group spike and slab variational bayes, 2023.

\bibitem[Kuss \& Rasmussen(2005)Kuss and Rasmussen]{Kuss2005}
Kuss, M. and Rasmussen, C.~E.
\newblock {Assessing approximate inference for binary gaussian process classification}.
\newblock \emph{Journal of Machine Learning Research}, 6:\penalty0 1679--1704, 2005.
\newblock ISSN 15337928.

\bibitem[Martens(2020)]{Martens2020}
Martens, J.
\newblock {New insights and perspectives on the natural gradient method}.
\newblock \emph{Journal of Machine Learning Research}, 21:\penalty0 1--76, 2020.
\newblock ISSN 15337928.

\bibitem[Paszke et~al.(2019)Paszke, Gross, Massa, Lerer, Bradbury, Chanan, Killeen, Lin, Gimelshein, Antiga, Desmaison, K{\"{o}}pf, Yang, DeVito, Raison, Tejani, Chilamkurthy, Steiner, Fang, Bai, and Chintala]{Paszke2019}
Paszke, A., Gross, S., Massa, F., Lerer, A., Bradbury, J., Chanan, G., Killeen, T., Lin, Z., Gimelshein, N., Antiga, L., Desmaison, A., K{\"{o}}pf, A., Yang, E., DeVito, Z., Raison, M., Tejani, A., Chilamkurthy, S., Steiner, B., Fang, L., Bai, J., and Chintala, S.
\newblock {PyTorch: An imperative style, high-performance deep learning library}.
\newblock \emph{Advances in Neural Information Processing Systems}, 32\penalty0 (NeurIPS), 2019.
\newblock ISSN 10495258.

\bibitem[Polson et~al.(2013)Polson, Scott, and Windle]{Polson2013}
Polson, N.~G., Scott, J.~G., and Windle, J.
\newblock {Bayesian inference for logistic models using P{\'{o}}lya-Gamma latent variables}.
\newblock \emph{Journal of the American Statistical Association}, 108\penalty0 (504):\penalty0 1339--1349, 2013.
\newblock ISSN 1537274X.
\newblock \doi{10.1080/01621459.2013.829001}.

\bibitem[Rasmussen \& Williams(2006)Rasmussen and Williams]{RasmussenW06}
Rasmussen, C.~E. and Williams, C. K.~I.
\newblock \emph{Gaussian processes for machine learning.}
\newblock Adaptive computation and machine learning. MIT Press, 2006.
\newblock ISBN 026218253X.

\bibitem[Ray et~al.(2020)Ray, Szabo, and Clara]{Ray2020}
Ray, K., Szabo, B., and Clara, G.
\newblock {Spike and slab variational Bayes for high dimensional logistic regression}.
\newblock In Larochelle, H., Ranzato, M., Hadsell, R., Balcan, M.~F., and Lin, H. (eds.), \emph{Advances in Neural Information Processing Systems}, volume~33, pp.\  14423--14434. Curran Associates, Inc., 2020.
\newblock URL \url{https://proceedings.neurips.cc/paper/2020/file/a5bad363fc47f424ddf5091c8471480a-Paper.pdf}.

\bibitem[Titsias(2009)]{Titsias2009}
Titsias, M.
\newblock Variational learning of inducing variables in sparse gaussian processes.
\newblock In van Dyk, D. and Welling, M. (eds.), \emph{Proceedings of the Twelth International Conference on Artificial Intelligence and Statistics}, volume~5 of \emph{Proceedings of Machine Learning Research}, pp.\  567--574. PMLR, 16--18 Apr 2009.
\newblock URL \url{https://proceedings.mlr.press/v5/titsias09a.html}.

\bibitem[Wang \& Pinar(2021)Wang and Pinar]{Wang2021}
Wang, F. and Pinar, A.
\newblock The multiple instance learning gaussian process probit model.
\newblock In Banerjee, A. and Fukumizu, K. (eds.), \emph{Proceedings of The 24th International Conference on Artificial Intelligence and Statistics}, volume 130 of \emph{Proceedings of Machine Learning Research}, pp.\  3034--3042. PMLR, 13--15 Apr 2021.
\newblock URL \url{https://proceedings.mlr.press/v130/wang21h.html}.

\bibitem[Wenzel et~al.(2017)Wenzel, Galy-Fajou, Donner, Kloft, and Opper]{Wenzel2017}
Wenzel, F., Galy-Fajou, T., Donner, C., Kloft, M., and Opper, M.
\newblock {Scalable Logit Gaussian Process Classification}.
\newblock \emph{Advances in Neural Information Processing Systems}, 30\penalty0 (NeurIPS), 2017.
\newblock URL \url{http://approximateinference.org/2017/accepted/WenzelEtAl2017.pdf}.

\bibitem[Zhan et~al.(2023)Zhan, Baise, and Moaveni1]{Zhan2023}
Zhan, W., Baise, L.~G., and Moaveni1, B.
\newblock {An Uncertainty Quantification Framework for Logistic Regression based Geospatial Natural Hazard Modeling}, 2023.

\bibitem[Zhang et~al.(2019{\natexlab{a}})Zhang, Butepage, Kjellstrom, and Mandt]{Zhang2019a}
Zhang, C., Butepage, J., Kjellstrom, H., and Mandt, S.
\newblock {Advances in Variational Inference}.
\newblock \emph{IEEE Transactions on Pattern Analysis and Machine Intelligence}, 41\penalty0 (8):\penalty0 2008--2026, 2019{\natexlab{a}}.
\newblock ISSN 19393539.
\newblock \doi{10.1109/TPAMI.2018.2889774}.

\bibitem[Zhang et~al.(2019{\natexlab{b}})Zhang, Dai, Kian, and Low]{Z2019}
Zhang, Y., Dai, Z., Kian, B., and Low, H.
\newblock {Bayesian Optimization with Binary Auxiliary Information}.
\newblock \emph{Proceedings of The 35th Uncertainty in Artificial Intelligence Conference}, 115:\penalty0 1222--1232, 2019{\natexlab{b}}.

\end{thebibliography}
\bibliographystyle{icml2024}

\newpage
\appendix
\onecolumn
\appendix

\section{Proofs} \label{appendix:proofs}

\subsection{Proof of Theorem \ref{thrm:nb}} \label{appendix:proof_bound}

\textit{Proof.} Write ${Z} = X - \vartheta$ and denote the density of ${Z}$ as $\phi_z$. It follows that,
{\allowdisplaybreaks %
\begin{align}
    \E_{X} &\ [  \log(1  + \exp(X)) ] = 
    \E_{Z} [ \log(1+ \exp(Z + \vartheta)) ]
    \nonumber
    \\
    =&\ 
	\int_{-\vartheta}^\infty \log(1 + \exp(z + \vartheta)) \phi_z dz
	+ \int_{-\infty}^{-\vartheta} \log(1 + \exp(z + \vartheta)) \phi_z dz
    \nonumber
    \\
    =&\ 
	\int_{-\vartheta}^\infty (z + \vartheta) \phi_z dz
	+ \int_{-\vartheta}^\infty \log[1 + \exp(-(z+\vartheta))] \phi_z dz 
    \nonumber \\
    +&\ \int_{-\infty}^{-\vartheta} \log(1 + \exp(z + \vartheta))\phi_z dz
    \nonumber
    \\
    \leq &\ 
	\int_{-\vartheta}^\infty (z + \vartheta) \phi_z dz
	+ \sum_{k=1}^{2l-1} \frac{-1^{k-1}}{k} \Big(
		\int_{-\vartheta}^\infty \exp(-k(z+\vartheta)) \phi_z dz 
    \nonumber
    \\
	+&\ \int_{-\infty}^{-\vartheta} \exp(k(z + \vartheta)) \phi_z dz
	\Big)
    \nonumber
\end{align}%
}%
where the inequality follows from the truncated Maclaurin series of $\log(1+x) \leq \sum_{k=1}^{2l-1} (-1)^{k-1}x^k/k$ for $x \in [0, 1]$, $l\geq 1$, and \eqref{eq:new_tight} follows from the fact that $\phi(z)' = - z\phi(z) / \tau^2 $ and $ \int_{a}^b e^{tz} \phi_z dz = e^{\tau^2 t^2 / 2} \left[ \Phi(b/\tau - t \tau) - \Phi(a/\tau - t\tau) \right]$. $\qed$

\subsection{Proof of \Cref{lemma:asymptotic}} \label{appendix:limit}

Here we study the limiting behavior of the terms in the sum of Theorem \ref{thrm:nb}. Recall, that the absolute value of the term is given by,
\begin{equation}
\begin{aligned}
    a_k = \frac{1}{k}  \Bigg[
        e^{k \vartheta}{}^{+\frac{k^2  \tau^2}{2}} \Phi \left(\! - \frac{\vartheta}{\tau}\! - \! k \tau \right) 
      + e^{-k \vartheta + \frac{k^2 \tau^2}{2}} \Phi \left( \frac{\vartheta}{\tau}\! -\!  k \tau \right)
    \Bigg].
\end{aligned}
\end{equation}
Using the fact that $\Phi(-t) \sim \frac{e^{-t^2/2}}{\sqrt{2\pi} t}$ as $t \to \infty$, we have,
\begin{align}
    a_k \sim &\
    \frac{1}{k} \Bigg[
            \exp{ \left( k \vartheta + k^2 \tau^2 / 2 \right) }  
            \frac{
                \exp{\left( -\frac{1}{2} \left( \frac{\vartheta}{\tau} + k \tau \right)^2 \right)}
            }{
                \sqrt{2\pi} \left( \frac{\vartheta}{\tau} + k \tau \right)
            }
        +
            \exp{\left( -k \vartheta + k^2 \tau^2 /2 \right)} 
            \frac{
                \exp{\left( -\frac{1}{2} \left( k \tau 
                - \frac{\vartheta}{\tau} 
                \right)^2 \right)}
            }{
                \sqrt{2\pi} \left( k \tau - \frac{\vartheta}{\tau} \right)
            }
    \Bigg]
    \nonumber
    \\ \nonumber
    = &\
    \frac{1}{k} \frac{\exp{\left( - \frac{\vartheta^2}{2 \tau^2} \right)}}{\sqrt{2\pi}}
    \Bigg[
        \frac{2k\tau}{k^2 \tau^4 - \vartheta^2}
    \Bigg]
    \\ \nonumber
    \sim &\ \frac{1}{k^2} 
\end{align}
$\qed$

\subsection{Proof of \Cref{cor:bound}} \label{appendix:proof_cor}

Let 
$$S_{K} = \frac{\tau}{\sqrt{2\pi}} e^{- \frac{\vartheta^2}{2 \tau^2}}+ \vartheta \Phi \left(\frac{\vartheta}{\tau} \right) + \sum_{k=1}^K (-1)^{(k-1)} a_k$$
where $a_k$ is the $k$th term in the sum of \eqref{eq:new_tight} as define above, then
\begin{equation}
    S_{2k} \leq \E_{X} \log(1 + \exp(X)) \leq S_{2k + 1}
\end{equation}
and so
\begin{equation}
    0 \leq \E_{X} \log(1 + \exp(X)) - S_{2k} \leq S_{2k +1} - S_{2k} = a_{2k + 1}
\end{equation}
Applying \Cref{lemma:asymptotic} and taking the limit as $k \to \infty$, we have
\begin{align}
    0 \leq \E_{X} \log(1 + \exp(X)) - S_{2k} \leq 0
\end{align}
and so $\lim_{k \to \infty} S_{2k} = \E_{X} \log(1 + \exp(X))$. $\qed$

\section{Additional Numerical Results} \label{appendix:additional_results}

\subsection{Error of Bounds} \label{appendix:error_of_bounds}

Here we present additional results for the error of the bounds. In particular, we compute the relative error of the bound by \citet{Jakkola97} and the proposed bound with $l=12$. Notably the relative error is computed with respect
to the Monte Carlo estimate of the expectation of $\log(1 + \exp(X))$ with $5 \times 10^6$ samples, and is given by the absolute difference between the bound and the ground truth, divided by the ground truth itself.
These results are presented in \Cref{fig:error_of_bound} and show that the proposed bound obtains a relative error that is smaller than that of the bound by \citet{Jakkola97}, particularly outside the origin of $\vartheta$ 
and $\tau$.

\begin{figure}[H]
\includegraphics[width=\textwidth]{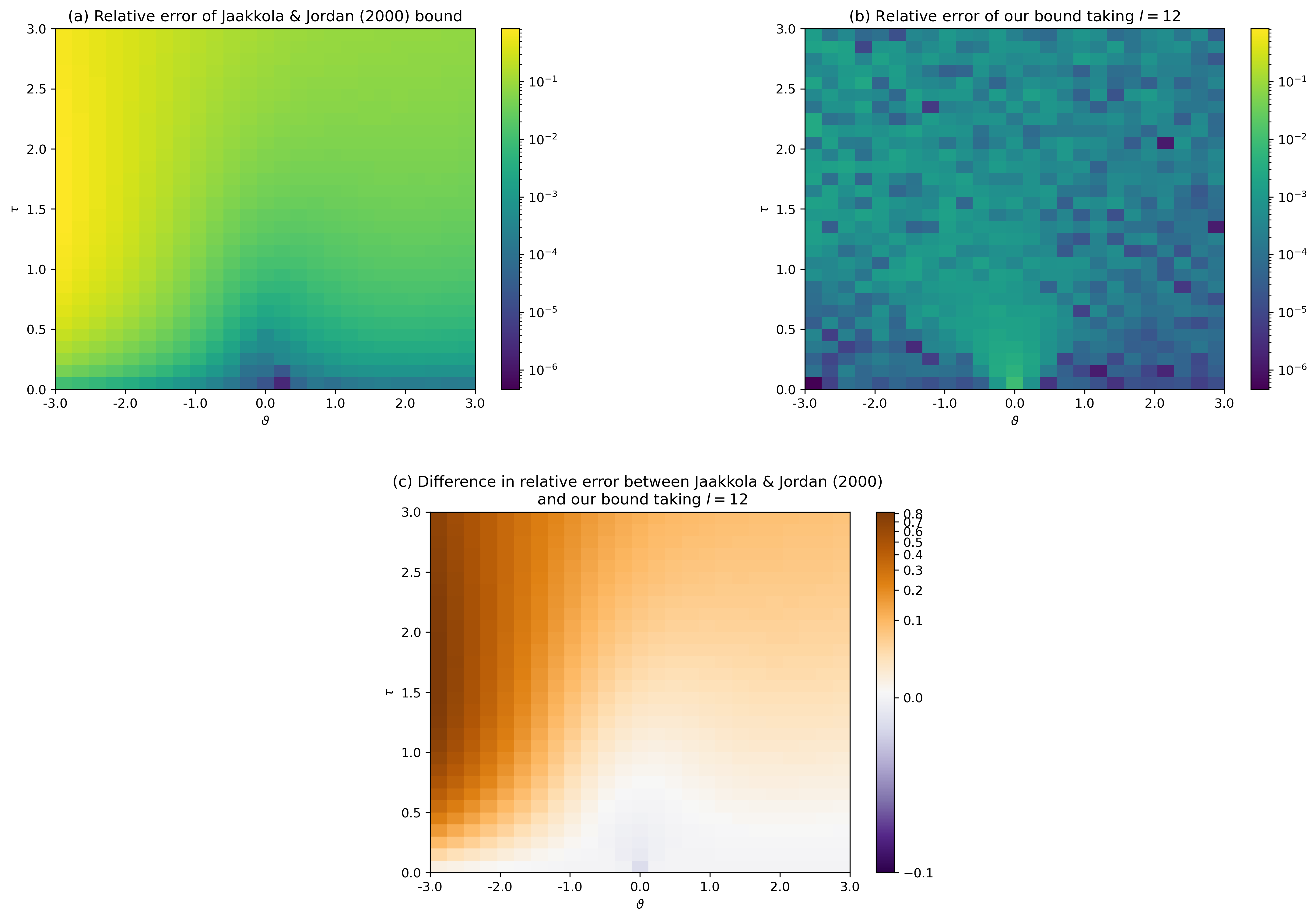}
\caption{Comparison of the relative error of the (a) \citet {Jakkola97} bound, (b) the proposed bound, (c) difference between the relative error of the bounds. The comparison is over a grid of values of $\vartheta$ and $\tau$.
Here the relative error of the bounds is the absolute difference between the bound and 
the ground truth, divided by the ground truth itself, where the ground truth is the 
expectation of $\log(1+\exp(X))$ computed using Monte Carlo with $5 \times 10^6$ samples. 
}
\label{fig:error_bounds}
\end{figure}

\subsection{Impact of $l$} \label{appendix:impact_of_l}

\begin{figure}[H]
\includegraphics[width=\textwidth]{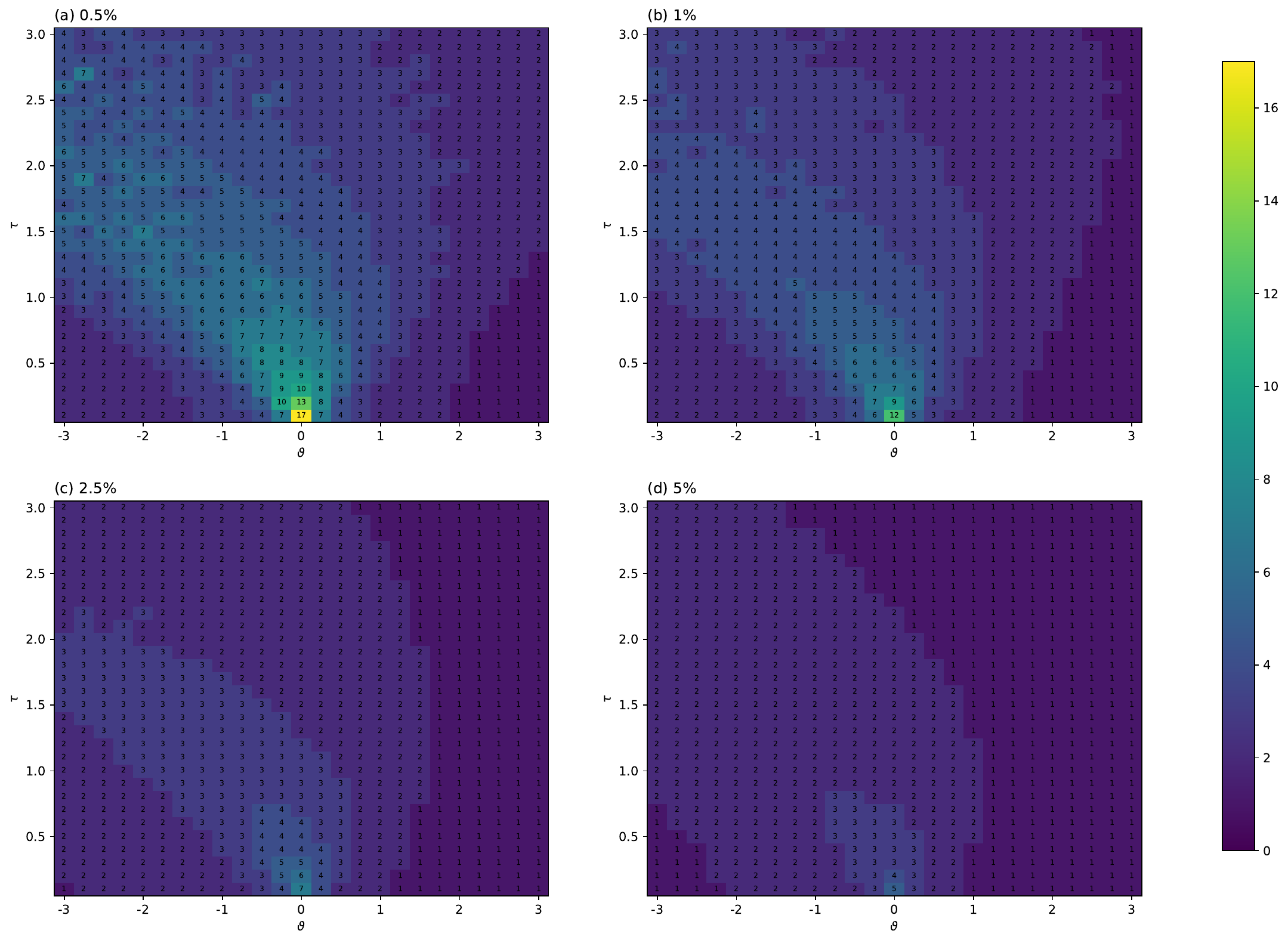}
\caption{The number of terms ($l$) needed such that the relative error is below (a) 0.5\%, (b) 1\%, (c) 2.5\% and (d) 5\% for different values of $\tau$ and $\vartheta$.}
\label{fig:l_terms}
\end{figure}

Here we present the values of $l$ need to obtain a relative error of less than 0.5\%, 1\%, 2.5\% and 5\% for different values of $\tau$ and $\vartheta$. These results are presented in \Cref{fig:l_terms} and show that the number of terms needed to obtain a relative error of less than 0.5\% is less than 17 for all values of $\tau$ and $\vartheta$ considered. Notably, this value decreases to 12, 7 and 5 for relative errors of less than 1\%, 2.5\% and 5\% respectively. 

\subsection{Logistic Regression Simulation Study} \label{appendix:logistic_regression}

Throughout this section we present additional results for the logistic regression simulation study presented in Section \ref{sec:logistic-regression}. In particular, we consider varying values of $n = \{500, 1000, 10,000\}$  and varying values of $p = \{5, 10, 25\}$. Furthermore, we consider different sampling schemes for the predictors, $x_i$s, which include:
\begin{description}
   \item[Setting 1] $x_i \overset{\text{iid}}{\sim} N(0_p, I_p)$.
   \item[Setting 2] $x_i \overset{\text{iid}}{\sim} N(0_p, \Sigma)$ where $\Sigma_{ij} = 0.3^{|i - j|}$ for $i,j=1,\dots,p$.
   \item[Setting 3] $x_i \overset{\text{iid}}{\sim} N(0_p, W^{-1})$ where $W \sim \text{Wishart}(p + 3, I_p)$.
\end{description}
These settings are chosen to highlight the performance of the different methods under different levels of correlation between the predictors. Notably, Setting 1 corresponds to the case where the predictors are independent, Setting 2 corresponds to the case where the predictors are mildly correlated and Setting 3 corresponds to the case where the predictors are strongly correlated.

The results summarized in Tables \ref{tab:logistic-reg-results-setting-1} -- \ref{tab:logistic-reg-results-setting-3} highlight that \propmethod is able to achieve similar performance to \mcmethod (considered the ground truth amongst the variational methods), while being significantly faster to compute. Furthermore, \propmethod is able to achieve similar predictive performance as with \pgmethod in terms of the AUC, however our method shows significant improvements in terms of the uncertainty quantification. This is made particularly evident as the coverage and CI widths are inline with \mcmethod whereas \pgmethod underestimates the posterior variance resulting in lower values for these quantities. Finally, the KL divergence between \mcmethod and \propmethod is significantly lower than that of \mcmethod and \pgmethod\!\!, meaning that \propmethod is in closer agreement with \mcmethod\!\!.

Furthermore, we note that the MSE, coverage and CI width are comparable to those of MCMC (considered the gold standard in Bayesian inference). This indicates that the variational posterior computed via \propmethod is an excellent approximation to the true posterior, whilst requiring an order of magnitude less computation time.

\section{Application to Real Data} \label{appendix:application}

\subsection{Soil Liquefaction Additional Results} \label{appendix:soil_liquefaction}
 
\begin{figure*}[ht]
    \centering
\makebox[\textwidth][c]{
\resizebox{1.0\textwidth}{!}{ %
    \includegraphics[width=1.00\textwidth]{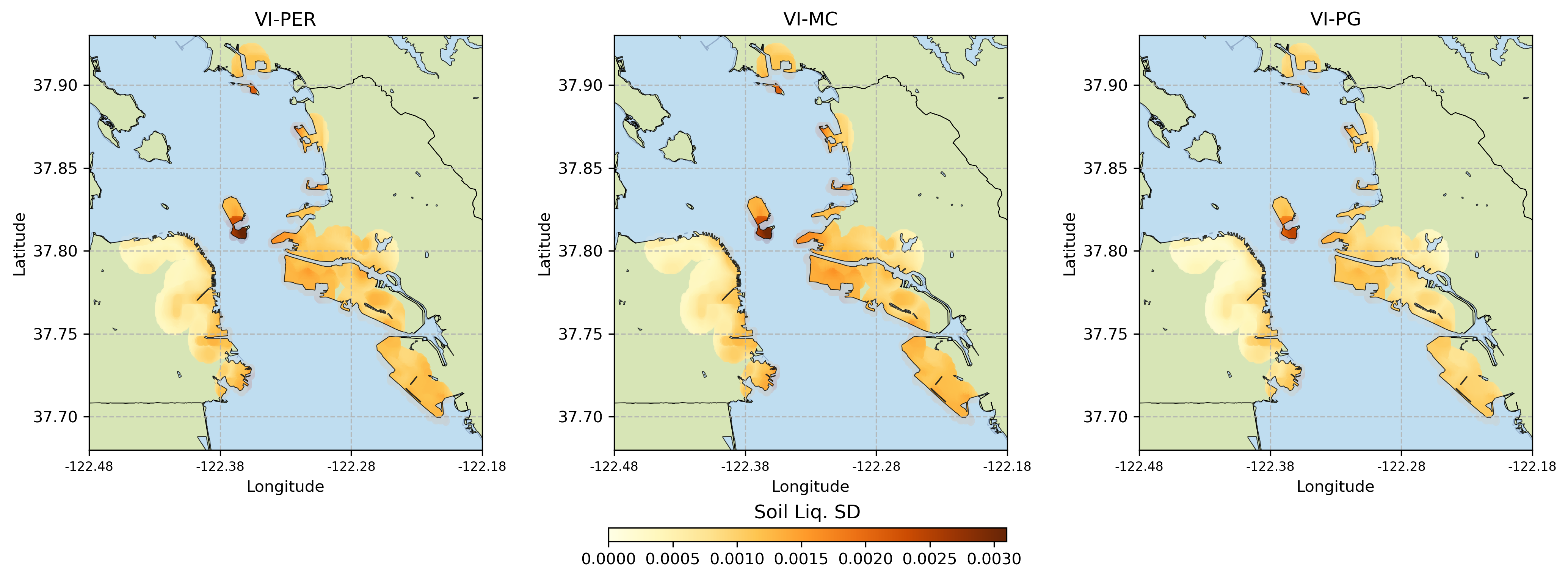} %
}}
    \vspace{-2em}
    \caption{\textit{Application to Soil liquefaction}. Standard deviation of soil liquefaction probability evaluated for the Loma Prieta earthquake for \propmethod, \mcmethod and \pgmethod under the variational family $\Q'$.}
    \label{fig:lomaprieta_full_Q}
\end{figure*}

Here we present additional results for the soil liquefaction application presented in Section \ref{sec:application}. In particular, \Cref{fig:lomaprieta_full_Q} shows the standard deviation of soil liquefaction probability evaluated for the Loma Prieta earthquake for \propmethod, \mcmethod and \pgmethod under the variational family $\Q'$. These results highlight that \propmethod propagates the uncertainty in the data inline with \mcmethod, whilst it appears \pgmethod underestimates this quantity as before.

\bgroup
\def\arraystretch{1.35}%
\begin{table}[h]
\centering
\caption{\textit{Logistic regression results, Setting 1:} Median (2.5\%, 97.5\% quantiles) of the ELBO, KL$_\text{MC}$, MSE, coverage, CI width and AUC for the different methods for data generated under Setting 1.}
\label{tab:logistic-reg-results-setting-1}
\vspace{1.0em}
\makebox[\textwidth][c]{
\resizebox{1.0\textwidth}{!}{ %
\begin{tabular}{| c | c c | c c c c c c l|}
    \hline
    Setting & VF & Method & ELBO & KL$_\text{MC}$ & MSE & Coverage & CI Width & AUC & \multicolumn{1}{c|}{Runtime} \\
    \hline 
    \multirow{7}{*}{500 / 5}
    & \multirow{3}{*}{Q}
    & \propmethod  & 
    -206 (-270, -160) & 0.00135 (0.0004, 0.0086) & 0.0873 (0.015, 0.39) & 0.988 (0.65, 1) & 1.13 (0.95, 1.3) & 0.908 (0.82, 0.95) & 6.5s (6.2s, 6.7s) \\
    && \mcmethod & 
    -206 (-270, -160) & - & 0.0873 (0.016, 0.38) & 0.982 (0.65, 1) & 1.12 (0.94, 1.3) & 0.908 (0.82, 0.95) & 50s (40s, 57s) \\
    && \pgmethod & 
    -217 (-290, -170) & 0.389 (0.14, 0.72) & 0.0908 (0.016, 0.37) & 0.87 (0.51, 1) & 0.879 (0.81, 0.95) & 0.908 (0.82, 0.95) & 0.02s (0.011s, 0.11s) \\
    \cline{2-3}
    & \multirow{3}{*}{Q'}
    & \propmethod & 
    -205 (-270, -160) & 0.0472 (0.02, 0.084) & 0.0858 (0.015, 0.39) & 1 (0.72, 1) & 1.24 (0.97, 1.5) & 0.908 (0.82, 0.95) & 1.2s (0.75s, 1.8s) \\
    && \mcmethod & 
    -205 (-270, -160) & - & 0.0871 (0.015, 0.39) & 1 (0.77, 1) & 1.29 (0.96, 1.6) & 0.908 (0.82, 0.95) & 50s (45s, 54s) \\
    && \pgmethod & 
    -206 (-270, -160) & 1.01 (0.23, 2) & 0.0896 (0.016, 0.37) & 0.862 (0.51, 1) & 0.873 (0.8, 0.95) & 0.908 (0.82, 0.95) & 0.088s (0.041s, 0.27s) \\
    \cline{2-3}
    & \multicolumn{2}{c|}{MCMC} & 
    - & - & 0.088 (0.015, 0.39) & 1 (0.71, 1) & 1.23 (0.96, 1.5) & 0.908 (0.82, 0.95) & 8m 21s (6m 16s, 9m 22s) \\
    \hline
    \multirow{7}{*}{500 / 10}
    & \multirow{3}{*}{Q}
    & \propmethod & 
    -173 (-210, -140) & 0.00219 (0.00095, 0.0048) & 0.242 (0.08, 0.86) & 0.946 (0.62, 1) & 1.78 (1.6, 2) & 0.947 (0.91, 0.97) & 5.6s (5.4s, 5.8s) \\
    && \mcmethod & 
    -173 (-210, -140) & - & 0.242 (0.081, 0.86) & 0.948 (0.62, 1) & 1.8 (1.6, 2) & 0.947 (0.91, 0.97) & 26s (22s, 30s) \\
    && \pgmethod & 
    -190 (-230, -150) & 1.17 (0.73, 1.7) & 0.233 (0.074, 0.93) & 0.828 (0.48, 1) & 1.34 (1.3, 1.4) & 0.947 (0.91, 0.97) & 0.021s (0.014s, 0.029s) \\
    \cline{2-3}
    & \multirow{3}{*}{Q'}
    & \propmethod & 
    -173 (-210, -140) & 0.111 (0.074, 0.17) & 0.246 (0.082, 0.85) & 0.964 (0.69, 1) & 1.96 (1.7, 2.3) & 0.947 (0.91, 0.97) & 0.59s (0.45s, 0.93s) \\
    && \mcmethod & 
    -173 (-210, -140) & - & 0.246 (0.08, 0.86) & 0.97 (0.72, 1) & 2.02 (1.7, 2.4) & 0.947 (0.91, 0.97) & 27s (24s, 30s) \\
    && \pgmethod & 
    -174 (-210, -140) & 2.49 (1.5, 4.1) & 0.233 (0.074, 0.93) & 0.834 (0.48, 1) & 1.34 (1.3, 1.4) & 0.947 (0.91, 0.97) & 0.084s (0.06s, 0.12s) \\
    \cline{2-3}
    & \multicolumn{2}{c|}{MCMC} & 
    - & - & 0.248 (0.08, 0.86) & 0.968 (0.71, 1) & 1.99 (1.7, 2.3) & 0.947 (0.91, 0.97) & 5m 16s (5m 11s, 5m 23s) \\
    \hline
    \multirow{7}{*}{500 / 25}
    & \multirow{3}{*}{Q}
    & \propmethod & 
    -150 (-180, -130) & 0.00876 (0.0046, 0.018) & 1.13 (0.47, 1.9) & 0.886 (0.75, 0.99) & 3.37 (3.1, 3.7) & 0.98 (0.96, 0.99) & 5.6s (5.3s, 6.2s) \\
    && \mcmethod & 
    -150 (-180, -130) & - & 1.13 (0.47, 1.9) & 0.886 (0.75, 0.99) & 3.39 (3.1, 3.7) & 0.98 (0.96, 0.99) & 45s (33s, 54s) \\
    && \pgmethod & 
    -176 (-210, -150) & 5.54 (4.1, 7.1) & 1.3 (0.5, 2.2) & 0.686 (0.56, 0.91) & 2.34 (2.2, 2.4) & 0.98 (0.96, 0.99) & 0.065s (0.044s, 0.092s) \\
    \cline{2-3}
    & \multirow{3}{*}{Q'}
    & \propmethod & 
    -149 (-180, -130) & 0.465 (0.38, 0.63) & 1.09 (0.48, 1.8) & 0.934 (0.82, 1) & 3.73 (3.3, 4.1) & 0.98 (0.96, 0.99) & 1.2s (0.88s, 2.3s) \\
    && \mcmethod & 
    -149 (-180, -130) & - & 1.09 (0.49, 1.8) & 0.946 (0.83, 1) & 3.89 (3.5, 4.3) & 0.98 (0.96, 0.99) & 46s (36s, 52s) \\
    && \pgmethod & 
    -152 (-180, -130) & 9.56 (6.9, 12) & 1.29 (0.5, 2.2) & 0.698 (0.56, 0.93) & 2.36 (2.2, 2.5) & 0.98 (0.96, 0.99) & 0.25s (0.18s, 0.39s) \\
    \cline{2-3}
    & \multicolumn{2}{c|}{MCMC} & 
    - & - & 1.09 (0.47, 1.8) & 0.942 (0.83, 1) & 3.76 (3.4, 4.2) & 0.98 (0.96, 0.99) & 8m 35s (6m 39s, 9m 59s) \\
    \hline
    \multirow{7}{*}{1000 / 5}
    & \multirow{3}{*}{Q}
    & \propmethod & 
    -391 (-550, -330) & 0.00108 (0.00026, 0.012) & 0.0427 (0.0081, 0.21) & 0.984 (0.61, 1) & 0.819 (0.65, 0.9) & 0.91 (0.81, 0.94) & 8.1s (7.4s, 8.7s) \\
    && \mcmethod & 
    -391 (-550, -330) & - & 0.042 (0.0082, 0.22) & 0.989 (0.61, 1) & 0.819 (0.66, 0.9) & 0.91 (0.81, 0.94) & 1m 49s (1m 18s, 2m 12s) \\
    && \pgmethod & 
    -417 (-590, -350) & 0.411 (0.11, 0.59) & 0.0424 (0.0076, 0.22) & 0.886 (0.47, 1) & 0.628 (0.57, 0.66) & 0.91 (0.81, 0.94) & 0.022s (0.011s, 0.032s) \\
    \cline{2-3}
    & \multirow{3}{*}{Q'}
    & \propmethod & 
    -391 (-550, -330) & 0.0348 (0.015, 0.086) & 0.0426 (0.0082, 0.21) & 1 (0.67, 1) & 0.921 (0.66, 1.1) & 0.91 (0.81, 0.94) & 1.5s (1s, 1.9s) \\
    && \mcmethod & 
    -391 (-550, -330) & - & 0.0428 (0.0083, 0.21) & 0.998 (0.63, 1) & 0.889 (0.65, 1.1) & 0.91 (0.81, 0.94) & 1m 40s (1m 27s, 1m 59s) \\
    && \pgmethod & 
    -391 (-550, -330) & 0.743 (0.12, 1.4) & 0.0421 (0.0078, 0.22) & 0.885 (0.47, 1) & 0.636 (0.57, 0.67) & 0.91 (0.81, 0.94) & 0.097s (0.049s, 0.14s) \\
    \cline{2-3}
    & \multicolumn{2}{c|}{MCMC} & 
    - & - & 0.0429 (0.0082, 0.21) & 0.998 (0.66, 1) & 0.9 (0.65, 1.1) & 0.91 (0.81, 0.94) & 8m 47s (6m 23s, 9m 28s) \\
    \hline
    \multirow{7}{*}{1000 / 10}
    & \multirow{3}{*}{Q}
    & \propmethod & 
    -325 (-400, -270) & 0.00128 (0.00057, 0.0073) & 0.126 (0.046, 0.59) & 0.93 (0.59, 1) & 1.28 (1.1, 1.4) & 0.946 (0.91, 0.96) & 5.7s (5.4s, 6.1s) \\
    && \mcmethod & 
    -325 (-400, -270) & - & 0.128 (0.046, 0.59) & 0.934 (0.59, 1) & 1.28 (1.1, 1.4) & 0.946 (0.91, 0.97) & 1m 4.8s (54s, 1m 17s) \\
    && \pgmethod & 
    -367 (-450, -310) & 1.12 (0.73, 1.6) & 0.131 (0.045, 0.63) & 0.808 (0.45, 0.99) & 0.96 (0.91, 1) & 0.946 (0.91, 0.96) & 0.026s (0.018s, 0.037s) \\
    \cline{2-3}
    & \multirow{3}{*}{Q'}
    & \propmethod & 
    -325 (-400, -270) & 0.141 (0.08, 0.2) & 0.127 (0.046, 0.57) & 0.973 (0.68, 1) & 1.46 (1.2, 1.7) & 0.946 (0.91, 0.96) & 0.94s (0.69s, 1.8s) \\
    && \mcmethod & 
    -325 (-400, -270) & - & 0.128 (0.046, 0.57) & 0.968 (0.67, 1) & 1.44 (1.2, 1.7) & 0.946 (0.91, 0.96) & 1m 4s (59s, 1m 13s) \\
    && \pgmethod & 
    -326 (-400, -270) & 2.22 (1.1, 4) & 0.129 (0.045, 0.63) & 0.809 (0.44, 0.98) & 0.961 (0.9, 1) & 0.946 (0.91, 0.96) & 0.11s (0.078s, 0.15s) \\
    \cline{2-3}
    & \multicolumn{2}{c|}{MCMC} & 
    - & - & 0.127 (0.046, 0.57) & 0.971 (0.69, 1) & 1.46 (1.2, 1.7) & 0.946 (0.91, 0.97) & 8m 32s (8m 21s, 8m 47s) \\
    \hline
    \multirow{7}{*}{1000 / 25}
    & \multirow{3}{*}{Q}
    & \propmethod & 
    -265 (-310, -230) & 0.00419 (0.002, 0.0088) & 0.496 (0.21, 1.2) & 0.918 (0.71, 0.99) & 2.41 (2.2, 2.6) & 0.977 (0.97, 0.98) & 7.1s (6.1s, 8s) \\
    && \mcmethod & 
    -265 (-310, -230) & - & 0.497 (0.21, 1.2) & 0.918 (0.71, 0.99) & 2.39 (2.2, 2.6) & 0.977 (0.97, 0.98) & 1m 31s (1m 8.2s, 1m 57s) \\
    && \pgmethod & 
    -336 (-390, -290) & 5.15 (4, 6.7) & 0.535 (0.22, 1.4) & 0.737 (0.51, 0.93) & 1.67 (1.6, 1.7) & 0.977 (0.97, 0.98) & 0.097s (0.07s, 0.14s) \\
    \cline{2-3}
    & \multirow{3}{*}{Q'}
    & \propmethod & 
    -264 (-310, -230) & 0.632 (0.53, 0.82) & 0.495 (0.21, 1.2) & 0.956 (0.78, 1) & 2.71 (2.4, 3) & 0.977 (0.97, 0.98) & 2.7s (1.5s, 4.3s) \\
    && \mcmethod & 
    -265 (-310, -230) & - & 0.493 (0.21, 1.1) & 0.952 (0.79, 1) & 2.7 (2.4, 3) & 0.977 (0.97, 0.98) & 1m 33s (1m 20s, 1m 44s) \\
    && \pgmethod & 
    -267 (-310, -230) & 8.42 (6, 12) & 0.531 (0.22, 1.4) & 0.745 (0.51, 0.93) & 1.68 (1.6, 1.8) & 0.977 (0.97, 0.98) & 0.42s (0.29s, 0.61s) \\
    \cline{2-3}
    & \multicolumn{2}{c|}{MCMC} & 
    - & - & 0.497 (0.21, 1.2) & 0.955 (0.81, 1) & 2.73 (2.5, 3) & 0.977 (0.97, 0.98) & 9m 13s (6m 57s, 10m 42s) \\
    \hline
    \multirow{7}{*}{10000 / 5}
    & \multirow{3}{*}{Q}
    & \propmethod & 
    -3920 (-4900, -3200) & 0.00815 (0.0052, 0.055) & 0.00536 (0.0008, 0.02) & 0.957 (0.63, 1) & 0.26 (0.23, 0.29) & 0.904 (0.85, 0.94) & 59s (47s, 1m 8.8s) \\
    && \mcmethod & 
    -3920 (-4900, -3200) & - & 0.00556 (0.00084, 0.021) & 0.953 (0.62, 1) & 0.259 (0.22, 0.29) & 0.904 (0.85, 0.94) & 20m 10s (4m 49s, 24m 14s) \\
    && \pgmethod & 
    -4270 (-5300, -3400) & 0.391 (0.17, 0.77) & 0.00552 (0.00086, 0.021) & 0.813 (0.48, 1) & 0.198 (0.19, 0.21) & 0.904 (0.85, 0.94) & 0.085s (0.048s, 0.15s) \\
    \cline{2-3}
    & \multirow{3}{*}{Q'}
    & \propmethod & 
    -3920 (-4900, -3200) & 0.0538 (0.023, 0.13) & 0.00574 (0.00086, 0.02) & 0.983 (0.71, 1) & 0.284 (0.23, 0.36) & 0.904 (0.85, 0.94) & 13s (9.6s, 18s) \\
    && \mcmethod & 
    -3920 (-4900, -3200) & - & 0.00553 (0.0008, 0.021) & 0.986 (0.71, 1) & 0.297 (0.24, 0.38) & 0.904 (0.85, 0.94) & 18m 24s (8m 17s, 22m 33s) \\
    && \pgmethod & 
    -3920 (-4900, -3200) & 1.05 (0.38, 2.3) & 0.00544 (0.00086, 0.021) & 0.817 (0.48, 1) & 0.201 (0.19, 0.22) & 0.904 (0.85, 0.94) & 0.42s (0.24s, 0.8s) \\
    \cline{2-3}
    & \multicolumn{2}{c|}{MCMC} & 
    - & - & 0.00566 (0.00084, 47) & 0.964 (0.045, 1) & 0.284 (0.084, 0.35) & 0.904 (0.85, 0.94) & 15m 11s (9m 54s, 18m 26s) \\
    \hline
    \multirow{7}{*}{10000 / 10}
    & \multirow{3}{*}{Q}
    & \propmethod & 
    -3060 (-3600, -2700) & 0.00712 (0.0055, 0.01) & 0.0124 (0.0043, 0.036) & 0.948 (0.72, 1) & 0.416 (0.38, 0.45) & 0.944 (0.92, 0.96) & 48s (42s, 55s) \\
    && \mcmethod & 
    -3060 (-3600, -2700) & - & 0.0121 (0.0045, 0.036) & 0.942 (0.71, 1) & 0.41 (0.37, 0.45) & 0.944 (0.92, 0.96) & 17m 39s (10m 13s, 18m 39s) \\
    && \pgmethod & 
    -3610 (-4300, -3200) & 1.1 (0.74, 1.5) & 0.0123 (0.0041, 0.036) & 0.823 (0.57, 0.99) & 0.305 (0.29, 0.32) & 0.944 (0.92, 0.96) & 0.23s (0.095s, 0.41s) \\
    \cline{2-3}
    & \multirow{3}{*}{Q'}
    & \propmethod & 
    -3060 (-3600, -2700) & 0.272 (0.13, 0.38) & 0.0124 (0.0043, 0.036) & 0.976 (0.83, 1) & 0.47 (0.41, 0.53) & 0.944 (0.92, 0.96) & 14s (9.8s, 25s) \\
    && \mcmethod & 
    -3060 (-3600, -2700) & - & 0.0124 (0.0042, 0.036) & 0.978 (0.85, 1) & 0.487 (0.41, 0.55) & 0.944 (0.92, 0.96) & 17m 30s (10m 14s, 18m 14s) \\
    && \pgmethod & 
    -3060 (-3600, -2700) & 3.06 (1.6, 4.5) & 0.0121 (0.0042, 0.035) & 0.837 (0.57, 0.99) & 0.308 (0.29, 0.32) & 0.944 (0.92, 0.96) & 1s (0.52s, 1.7s) \\
    \cline{2-3}
    & \multicolumn{2}{c|}{MCMC} & 
    - & - & 0.0123 (0.0043, 0.048) & 0.974 (0.76, 1) & 0.45 (0.4, 0.54) & 0.944 (0.92, 0.96) & 13m 60s (13m 35s, 14m 21s) \\
    \hline
    \multirow{7}{*}{10000 / 25}
    & \multirow{3}{*}{Q}
    & \propmethod & 
    -2140 (-2600, -1900) & 0.00603 (0.00076, 0.01) & 0.0518 (0.024, 0.15) & 0.916 (0.68, 0.99) & 0.784 (0.72, 0.84) & 0.974 (0.96, 0.98) & 30s (21s, 49s) \\
    && \mcmethod & 
    -2140 (-2600, -1900) & - & 0.0514 (0.024, 0.15) & 0.913 (0.68, 0.99) & 0.783 (0.71, 0.84) & 0.974 (0.96, 0.98) & 15m 2.9s (11m 43s, 18m 38s) \\
    && \pgmethod & 
    -3100 (-3700, -2700) & 4.79 (3.7, 5.8) & 0.0503 (0.024, 0.16) & 0.756 (0.49, 0.9) & 0.538 (0.51, 0.56) & 0.974 (0.96, 0.98) & 1s (0.51s, 2.5s) \\
    \cline{2-3}
    & \multirow{3}{*}{Q'}
    & \propmethod & 
    -2140 (-2600, -1900) & 1.61 (1.3, 2) & 0.0519 (0.024, 0.14) & 0.96 (0.8, 1) & 0.906 (0.79, 1) & 0.974 (0.96, 0.98) & 23s (11s, 49s) \\
    && \mcmethod & 
    -2140 (-2600, -1900) & - & 0.052 (0.024, 0.15) & 0.973 (0.83, 1) & 0.985 (0.83, 1.2) & 0.974 (0.96, 0.98) & 14m 8.7s (12m 0.36s, 16m 5.3s) \\
    && \pgmethod & 
    -2140 (-2600, -1900) & 13.7 (8.9, 22) & 0.0508 (0.024, 0.16) & 0.762 (0.49, 0.9) & 0.543 (0.51, 0.57) & 0.974 (0.96, 0.98) & 4.4s (2.2s, 11s) \\
    \cline{2-3}
    & \multicolumn{2}{c|}{MCMC} & 
    - & - & 0.0521 (0.025, 0.15) & 0.954 (0.78, 0.99) & 0.893 (0.79, 1) & 0.974 (0.96, 0.98) & 15m 6.9s (12m 33s, 20m 3.4s) \\
    \hline
\end{tabular}%
}}
\end{table}
\egroup

\bgroup
\def\arraystretch{1.35}%
\begin{table}[h]
\centering
\caption{\textit{Logistic regression results, Setting 2:} Median (2.5\%, 97.5\% quantiles) of the ELBO, KL$_\text{MC}$, MSE, coverage, CI width and AUC for the different methods for data generated under Setting 2.}
\label{tab:logistic-reg-results-setting-2}
\vspace{1.0em}
\makebox[\textwidth][c]{
\resizebox{1.0\textwidth}{!}{ %
\begin{tabular}{| c | c c | c c c c c c l|}
    \hline
    Setting & VF & Method & ELBO & KL$_\text{MC}$ & MSE & Coverage & CI Width & AUC & \multicolumn{1}{c|}{Runtime} \\
    \hline 
    \multirow{7}{*}{500 / 5}
    & \multirow{3}{*}{Q}
    & \propmethod  & 
    -207 (-280, -140) & 0.0013 (0.00043, 0.012) & 0.0876 (0.015, 0.47) & 0.972 (0.61, 1) & 1.15 (0.91, 1.5) & 0.907 (0.8, 0.96) & 6.8s (6.6s, 7.2s) \\
    && \mcmethod & 
    -207 (-280, -140) & - & 0.0861 (0.013, 0.47) & 0.966 (0.61, 1) & 1.14 (0.9, 1.5) & 0.907 (0.8, 0.96) & 54s (28s, 60s) \\
    && \pgmethod & 
    -217 (-300, -150) & 0.444 (0.1, 1.1) & 0.0875 (0.013, 0.46) & 0.856 (0.5, 1) & 0.882 (0.8, 1) & 0.907 (0.8, 0.96) & 0.021s (0.0096s, 0.05s) \\
    \cline{2-3}
    & \multirow{3}{*}{Q'}
    & \propmethod & 
    -206 (-280, -140) & 0.0362 (0.015, 0.076) & 0.0832 (0.014, 0.47) & 0.998 (0.7, 1) & 1.24 (0.93, 1.7) & 0.907 (0.8, 0.96) & 1.2s (0.79s, 1.6s) \\
    && \mcmethod & 
    -206 (-280, -140) & - & 0.0849 (0.012, 0.47) & 1 (0.74, 1) & 1.28 (0.95, 1.8) & 0.907 (0.8, 0.96) & 53s (30s, 58s) \\
    && \pgmethod & 
    -207 (-280, -140) & 0.897 (0.24, 2.6) & 0.0867 (0.013, 0.46) & 0.86 (0.49, 1) & 0.873 (0.79, 0.98) & 0.907 (0.8, 0.96) & 0.088s (0.039s, 0.19s) \\
    \cline{2-3}
    & \multicolumn{2}{c|}{MCMC} & 
    - & - & 0.0857 (0.012, 0.47) & 0.996 (0.69, 1) & 1.23 (0.92, 1.7) & 0.907 (0.8, 0.96) & 8m 9.2s (7m 6s, 8m 25s) \\
    \hline
    \multirow{7}{*}{500 / 10}
    & \multirow{3}{*}{Q}
    & \propmethod  & 
    -174 (-220, -130) & 0.00229 (0.00072, 0.0061) & 0.249 (0.069, 0.65) & 0.932 (0.74, 1) & 1.79 (1.6, 2.1) & 0.947 (0.91, 0.97) & 6.7s (6.1s, 7.1s) \\
    && \mcmethod & 
    -174 (-220, -130) & - & 0.25 (0.068, 0.66) & 0.936 (0.75, 1) & 1.81 (1.6, 2.1) & 0.947 (0.91, 0.97) & 45s (29s, 49s) \\
    && \pgmethod & 
    -191 (-240, -140) & 1.19 (0.68, 1.9) & 0.252 (0.061, 0.72) & 0.82 (0.54, 1) & 1.34 (1.3, 1.5) & 0.947 (0.91, 0.97) & 0.021s (0.013s, 0.074s) \\
    \cline{2-3}
    & \multirow{3}{*}{Q'}
    & \propmethod & 
    -173 (-220, -130) & 0.106 (0.067, 0.16) & 0.249 (0.073, 0.64) & 0.97 (0.81, 1) & 1.97 (1.7, 2.4) & 0.947 (0.91, 0.97) & 0.77s (0.47s, 1.3s) \\
    && \mcmethod & 
    -173 (-220, -130) & - & 0.249 (0.073, 0.65) & 0.98 (0.83, 1) & 2.03 (1.7, 2.6) & 0.947 (0.91, 0.97) & 47s (32s, 50s) \\
    && \pgmethod & 
    -174 (-220, -130) & 2.55 (1.4, 5) & 0.252 (0.061, 0.71) & 0.83 (0.55, 1) & 1.35 (1.3, 1.5) & 0.947 (0.91, 0.97) & 0.09s (0.056s, 0.23s) \\
    \cline{2-3}
    & \multicolumn{2}{c|}{MCMC} & 
    - & - & 0.243 (0.074, 0.65) & 0.976 (0.83, 1) & 2 (1.7, 2.5) & 0.947 (0.91, 0.97) & 7m 45s (7m 6.6s, 7m 56s) \\
    \hline
    \multirow{7}{*}{500 / 25}
    & \multirow{3}{*}{Q}
    & \propmethod  & 
    -152 (-180, -130) & 0.00846 (0.0048, 0.017) & 0.973 (0.57, 2.2) & 0.908 (0.73, 0.98) & 3.38 (3.1, 3.7) & 0.981 (0.97, 0.99) & 9.1s (8.8s, 9.7s) \\
    && \mcmethod & 
    -151 (-180, -130) & - & 0.982 (0.57, 2.2) & 0.91 (0.73, 0.98) & 3.4 (3.1, 3.7) & 0.98 (0.97, 0.99) & 43s (32s, 49s) \\
    && \pgmethod & 
    -177 (-210, -150) & 5.57 (4.3, 7.3) & 1.17 (0.55, 2.6) & 0.718 (0.54, 0.88) & 2.34 (2.2, 2.5) & 0.981 (0.97, 0.99) & 0.057s (0.04s, 0.081s) \\
    \cline{2-3}
    & \multirow{3}{*}{Q'}
    & \propmethod & 
    -149 (-170, -130) & 0.435 (0.35, 0.53) & 0.95 (0.56, 2.1) & 0.95 (0.78, 0.99) & 3.72 (3.3, 4.1) & 0.981 (0.97, 0.99) & 1.4s (0.88s, 2.1s) \\
    && \mcmethod & 
    -149 (-170, -130) & - & 0.952 (0.56, 2.1) & 0.958 (0.8, 0.99) & 3.83 (3.5, 4.4) & 0.98 (0.97, 0.99) & 46s (36s, 49s) \\
    && \pgmethod & 
    -152 (-180, -130) & 9.05 (6.9, 13) & 1.16 (0.55, 2.5) & 0.724 (0.54, 0.89) & 2.36 (2.3, 2.5) & 0.981 (0.97, 0.99) & 0.24s (0.17s, 0.32s) \\
    \cline{2-3}
    & \multicolumn{2}{c|}{MCMC} & 
    - & - & 0.953 (0.56, 2.1) & 0.954 (0.8, 0.99) & 3.76 (3.4, 4.2) & 0.981 (0.97, 0.99) & 7m 40s (6m 13s, 8m 42s) \\
    \hline
    \multirow{7}{*}{1000 / 5}
    & \multirow{3}{*}{Q}
    & \propmethod  & 
    -410 (-560, -310) & 0.00142 (0.00039, 0.012) & 0.0465 (0.011, 0.2) & 0.962 (0.63, 1) & 0.795 (0.63, 1) & 0.902 (0.79, 0.95) & 4.7s (4.4s, 5.1s) \\
    && \mcmethod & 
    -410 (-560, -310) & - & 0.0472 (0.011, 0.2) & 0.961 (0.62, 1) & 0.795 (0.64, 1) & 0.902 (0.79, 0.95) & 1m 7.6s (56s, 1m 43s) \\
    && \pgmethod & 
    -438 (-610, -320) & 0.379 (0.092, 0.97) & 0.0474 (0.011, 0.2) & 0.833 (0.47, 1) & 0.616 (0.56, 0.68) & 0.902 (0.79, 0.95) & 0.016s (0.0068s, 0.034s) \\
    \cline{2-3}
    & \multirow{3}{*}{Q'}
    & \propmethod & 
    -410 (-560, -300) & 0.0232 (0.0093, 0.074) & 0.0461 (0.011, 0.2) & 0.997 (0.69, 1) & 0.866 (0.65, 1.2) & 0.902 (0.79, 0.95) & 1s (0.7s, 4.4s) \\
    && \mcmethod & 
    -410 (-560, -300) & - & 0.0469 (0.011, 0.2) & 0.997 (0.67, 1) & 0.867 (0.64, 1.2) & 0.902 (0.79, 0.95) & 1m 4.7s (55s, 1m 15s) \\
    && \pgmethod & 
    -410 (-560, -310) & 0.684 (0.12, 2) & 0.0479 (0.011, 0.2) & 0.861 (0.48, 1) & 0.624 (0.56, 0.69) & 0.902 (0.79, 0.95) & 0.073s (0.031s, 0.17s) \\
    \cline{2-3}
    & \multicolumn{2}{c|}{MCMC} & 
    - & - & 0.0466 (0.011, 0.2) & 0.995 (0.66, 1) & 0.869 (0.63, 1.1) & 0.902 (0.79, 0.95) & 5m 47s (5m 40s, 7m 18s) \\
    \hline
    \multirow{7}{*}{1000 / 10}
    & \multirow{3}{*}{Q}
    & \propmethod  & 
    -332 (-420, -260) & 0.0013 (0.00057, 0.0086) & 0.125 (0.035, 0.43) & 0.94 (0.65, 1) & 1.27 (1.1, 1.4) & 0.944 (0.91, 0.97) & 4.6s (4.3s, 5.1s) \\
    && \mcmethod & 
    -332 (-420, -260) & - & 0.125 (0.036, 0.44) & 0.94 (0.65, 1) & 1.28 (1.1, 1.5) & 0.944 (0.91, 0.97) & 1m 3.6s (53s, 1m 17s) \\
    && \pgmethod & 
    -375 (-480, -290) & 1.14 (0.67, 1.7) & 0.128 (0.033, 0.47) & 0.806 (0.5, 1) & 0.953 (0.89, 1) & 0.944 (0.91, 0.97) & 0.026s (0.017s, 0.04s) \\
    \cline{2-3}
    & \multirow{3}{*}{Q'}
    & \propmethod & 
    -331 (-420, -260) & 0.112 (0.077, 0.18) & 0.119 (0.037, 0.43) & 0.982 (0.73, 1) & 1.44 (1.2, 1.8) & 0.944 (0.91, 0.97) & 0.97s (0.69s, 2.1s) \\
    && \mcmethod & 
    -331 (-420, -260) & - & 0.123 (0.037, 0.43) & 0.974 (0.73, 1) & 1.42 (1.2, 1.8) & 0.944 (0.91, 0.97) & 1m 2.4s (52s, 1m 7.2s) \\
    && \pgmethod & 
    -332 (-420, -260) & 2.2 (1, 4.6) & 0.128 (0.033, 0.47) & 0.819 (0.5, 1) & 0.956 (0.89, 1) & 0.944 (0.91, 0.97) & 0.11s (0.075s, 0.18s) \\
    \cline{2-3}
    & \multicolumn{2}{c|}{MCMC} & 
    - & - & 0.122 (0.037, 0.43) & 0.979 (0.75, 1) & 1.45 (1.2, 1.8) & 0.944 (0.91, 0.97) & 5m 30s (5m 23s, 5m 39s) \\
    \hline
    \multirow{7}{*}{1000 / 25}
    & \multirow{3}{*}{Q}
    & \propmethod  & 
    -267 (-320, -230) & 0.00359 (0.002, 0.0084) & 0.491 (0.23, 1.4) & 0.919 (0.69, 0.99) & 2.4 (2.2, 2.6) & 0.977 (0.96, 0.99) & 6.3s (4.8s, 7.6s) \\
    && \mcmethod & 
    -267 (-320, -230) & - & 0.483 (0.23, 1.4) & 0.913 (0.69, 0.99) & 2.4 (2.1, 2.6) & 0.977 (0.96, 0.99) & 2m 6.2s (57s, 2m 20s) \\
    && \pgmethod & 
    -339 (-410, -280) & 5.37 (3.7, 6.9) & 0.533 (0.27, 1.6) & 0.736 (0.48, 0.9) & 1.66 (1.6, 1.7) & 0.977 (0.96, 0.99) & 0.1s (0.045s, 0.19s) \\
    \cline{2-3}
    & \multirow{3}{*}{Q'}
    & \propmethod & 
    -265 (-320, -220) & 0.599 (0.48, 0.75) & 0.48 (0.23, 1.4) & 0.953 (0.77, 1) & 2.68 (2.3, 3) & 0.977 (0.96, 0.99) & 3s (1.8s, 5.1s) \\
    && \mcmethod & 
    -266 (-320, -230) & - & 0.476 (0.22, 1.3) & 0.951 (0.78, 1) & 2.67 (2.3, 3.1) & 0.977 (0.96, 0.99) & 2m 4.1s (58s, 2m 16s) \\
    && \pgmethod & 
    -268 (-320, -230) & 8.16 (5.2, 12) & 0.532 (0.26, 1.6) & 0.749 (0.49, 0.9) & 1.68 (1.6, 1.8) & 0.977 (0.96, 0.99) & 0.43s (0.19s, 0.77s) \\
    \cline{2-3}
    & \multicolumn{2}{c|}{MCMC} & 
    - & - & 0.498 (0.23, 1.3) & 0.957 (0.77, 1) & 2.69 (2.3, 3) & 0.977 (0.96, 0.99) & 6m 27s (6m 11s, 6m 42s) \\
    \hline
    \multirow{7}{*}{10000 / 5}
    & \multirow{3}{*}{Q}
    & \propmethod  & 
    -3880 (-5100, -2900) & 0.00891 (0.0048, 0.2) & 0.0052 (0.001, 0.026) & 0.961 (0.59, 1) & 0.264 (0.22, 0.33) & 0.906 (0.82, 0.95) & 37s (32s, 48s) \\
    && \mcmethod & 
    -3880 (-5100, -2900) & - & 0.00543 (0.001, 0.027) & 0.955 (0.57, 1) & 0.261 (0.21, 0.34) & 0.906 (0.82, 0.95) & 13m 46s (3m 46s, 15m 20s) \\
    && \pgmethod & 
    -4220 (-5600, -3200) & 0.424 (0.13, 1) & 0.00544 (0.00096, 0.027) & 0.806 (0.42, 1) & 0.198 (0.18, 0.22) & 0.906 (0.82, 0.95) & 0.05s (0.027s, 0.082s) \\
    \cline{2-3}
    & \multirow{3}{*}{Q'}
    & \propmethod & 
    -3880 (-5100, -2900) & 0.0414 (0.015, 1.2) & 0.00549 (0.0011, 0.027) & 0.983 (0.66, 1) & 0.288 (0.22, 0.38) & 0.906 (0.82, 0.95) & 9.2s (6.2s, 16s) \\
    && \mcmethod & 
    -3880 (-5100, -2900) & - & 0.00516 (0.00093, 0.027) & 0.993 (0.65, 1) & 0.296 (0.23, 0.4) & 0.906 (0.82, 0.95) & 12m 50s (3m 39s, 14m 0.56s) \\
    && \pgmethod & 
    -3880 (-5100, -2900) & 0.982 (0.3, 3) & 0.00538 (0.00097, 0.027) & 0.839 (0.43, 1) & 0.202 (0.18, 0.22) & 0.906 (0.82, 0.95) & 0.22s (0.12s, 0.4s) \\
    \cline{2-3}
    & \multicolumn{2}{c|}{MCMC} & 
    - & - & 0.00561 (0.001, 91) & 0.962 (0, 1) & 0.275 (0, 0.38) & 0.905 (0.82, 0.95) & 8m 6s (7m 55s, 8m 43s) \\
    \hline
    \multirow{7}{*}{10000 / 10}
    & \multirow{3}{*}{Q}
    & \propmethod  & 
    -3110 (-3700, -2500) & 0.00743 (0.0049, 0.013) & 0.0123 (0.0033, 0.041) & 0.948 (0.71, 1) & 0.415 (0.37, 0.47) & 0.942 (0.91, 0.96) & 35s (30s, 44s) \\
    && \mcmethod & 
    -3110 (-3700, -2500) & - & 0.0126 (0.0032, 0.04) & 0.937 (0.71, 1) & 0.409 (0.36, 0.47) & 0.942 (0.91, 0.96) & 14m 37s (12m 20s, 15m 40s) \\
    && \pgmethod & 
    -3660 (-4400, -2900) & 1.09 (0.68, 1.7) & 0.0124 (0.0029, 0.039) & 0.822 (0.55, 1) & 0.303 (0.29, 0.33) & 0.942 (0.91, 0.96) & 0.14s (0.078s, 0.31s) \\
    \cline{2-3}
    & \multirow{3}{*}{Q'}
    & \propmethod & 
    -3110 (-3700, -2500) & 0.155 (0.043, 0.26) & 0.0128 (0.0032, 0.043) & 0.976 (0.81, 1) & 0.464 (0.4, 0.56) & 0.942 (0.91, 0.96) & 11s (7.2s, 34s) \\
    && \mcmethod & 
    -3110 (-3700, -2500) & - & 0.0125 (0.0031, 0.04) & 0.979 (0.83, 1) & 0.474 (0.4, 0.6) & 0.942 (0.91, 0.96) & 13m 23s (2m 37s, 14m 23s) \\
    && \pgmethod & 
    -3110 (-3700, -2500) & 2.75 (1.5, 5.2) & 0.0123 (0.0029, 0.039) & 0.829 (0.56, 1) & 0.307 (0.29, 0.33) & 0.942 (0.91, 0.96) & 0.68s (0.43s, 1.3s) \\
    \cline{2-3}
    & \multicolumn{2}{c|}{MCMC} & 
    - & - & 0.0125 (0.0031, 0.04) & 0.967 (0.77, 1) & 0.446 (0.39, 0.55) & 0.942 (0.91, 0.96) & 9m 7.4s (8m 58s, 9m 45s) \\
    \hline
    \multirow{7}{*}{10000 / 25}
    & \multirow{3}{*}{Q}
    & \propmethod  & 
    -2160 (-2600, -1900) & 0.00512 (0.00074, 0.011) & 0.0523 (0.02, 0.18) & 0.912 (0.62, 1) & 0.782 (0.71, 0.85) & 0.974 (0.96, 0.98) & 28s (25s, 39s) \\
    && \mcmethod & 
    -2160 (-2600, -1900) & - & 0.0523 (0.02, 0.19) & 0.913 (0.62, 1) & 0.783 (0.71, 0.85) & 0.974 (0.96, 0.98) & 13m 57s (13m 4.5s, 15m 34s) \\
    && \pgmethod & 
    -3120 (-3700, -2700) & 4.78 (3.7, 6) & 0.0537 (0.021, 0.2) & 0.744 (0.44, 0.94) & 0.537 (0.51, 0.56) & 0.974 (0.96, 0.98) & 0.93s (0.64s, 1.6s) \\
    \cline{2-3}
    & \multirow{3}{*}{Q'}
    & \propmethod & 
    -2150 (-2600, -1900) & 1.39 (0.67, 1.8) & 0.0527 (0.02, 0.18) & 0.953 (0.71, 1) & 0.9 (0.78, 1) & 0.974 (0.96, 0.98) & 23s (11s, 45s) \\
    && \mcmethod & 
    -2160 (-2600, -1900) & - & 0.0527 (0.02, 0.19) & 0.967 (0.78, 1) & 0.97 (0.84, 1.2) & 0.974 (0.96, 0.98) & 13m 31s (6m 5s, 14m 34s) \\
    && \pgmethod & 
    -2160 (-2600, -1900) & 13 (8.9, 20) & 0.0539 (0.021, 0.2) & 0.752 (0.44, 0.95) & 0.541 (0.51, 0.57) & 0.974 (0.96, 0.98) & 4.1s (2.7s, 6.6s) \\
    \cline{2-3}
    & \multicolumn{2}{c|}{MCMC} & 
    - & - & 0.0528 (0.022, 0.19) & 0.952 (0.72, 1) & 0.9 (0.78, 1) & 0.974 (0.96, 0.98) & 11m 31s (11m 16s, 12m 1.2s) \\
    \hline
\end{tabular}%
}}
\end{table}
\egroup

\bgroup
\def\arraystretch{1.35}%
\begin{table}[h]
\centering
\caption{\textit{Logistic regression results, Setting 3:} Median (2.5\%, 97.5\% quantiles) of the ELBO, KL$_\text{MC}$, MSE, coverage, CI width and AUC for the different methods for data generated under Setting 3.}
\label{tab:logistic-reg-results-setting-3}
\vspace{1.0em}
\makebox[\textwidth][c]{
\resizebox{1.0\textwidth}{!}{ %
\begin{tabular}{| c | c c | c c c c c c l|}
    \hline
    Setting & VF & Method & ELBO & KL$_\text{MC}$ & MSE & Coverage & CI Width & AUC & \multicolumn{1}{c|}{Runtime} \\
    \hline 
    \multirow{7}{*}{500 / 5}
    & \multirow{3}{*}{Q}
    & \propmethod  & 
    -209 (-280, -150) & 0.00154 (0.0004, 0.01) & 0.0871 (0.018, 0.33) & 0.972 (0.67, 1) & 1.13 (0.89, 1.4) & 0.903 (0.81, 0.95) & 4s (3.8s, 4.4s) \\
    && \mcmethod & 
    -209 (-280, -150) & - & 0.0846 (0.017, 0.33) & 0.964 (0.67, 1) & 1.13 (0.89, 1.4) & 0.903 (0.81, 0.95) & 25s (21s, 30s) \\
    && \pgmethod & 
    -219 (-300, -160) & 0.428 (0.1, 1.1) & 0.0872 (0.019, 0.33) & 0.856 (0.51, 1) & 0.871 (0.78, 0.98) & 0.903 (0.81, 0.95) & 0.012s (0.0059s, 0.024s) \\
    \cline{2-3}
    & \multirow{3}{*}{Q'}
    & \propmethod & 
    -209 (-280, -150) & 0.039 (0.015, 0.081) & 0.0807 (0.018, 0.33) & 1 (0.7, 1) & 1.21 (0.93, 1.7) & 0.903 (0.81, 0.95) & 0.52s (0.37s, 1.4s) \\
    && \mcmethod & 
    -208 (-280, -150) & - & 0.0844 (0.017, 0.32) & 1 (0.73, 1) & 1.24 (0.94, 1.7) & 0.903 (0.81, 0.95) & 26s (23s, 28s) \\
    && \pgmethod & 
    -209 (-280, -150) & 0.867 (0.18, 2.2) & 0.0873 (0.018, 0.33) & 0.862 (0.51, 1) & 0.861 (0.79, 0.96) & 0.903 (0.81, 0.95) & 0.051s (0.024s, 0.11s) \\
    \cline{2-3}
    & \multicolumn{2}{c|}{MCMC} & 
    - & - & 0.0849 (0.019, 0.32) & 1 (0.71, 1) & 1.2 (0.92, 1.7) & 0.903 (0.81, 0.95) & 5m 6.3s (5m 1.2s, 5m 12s) \\
    \hline
    \multirow{7}{*}{500 / 10}
    & \multirow{3}{*}{Q}
    & \propmethod  & 
    -174 (-230, -130) & 0.0024 (0.0012, 0.006) & 0.242 (0.091, 1.1) & 0.926 (0.64, 1) & 1.81 (1.5, 2.2) & 0.946 (0.89, 0.97) & 4.5s (3.8s, 5.5s) \\
    && \mcmethod & 
    -174 (-230, -130) & - & 0.244 (0.091, 1.1) & 0.926 (0.65, 1) & 1.83 (1.5, 2.2) & 0.946 (0.89, 0.97) & 33s (24s, 50s) \\
    && \pgmethod & 
    -190 (-260, -140) & 1.32 (0.54, 2.4) & 0.251 (0.084, 1.1) & 0.794 (0.5, 0.99) & 1.34 (1.2, 1.5) & 0.946 (0.89, 0.97) & 0.028s (0.015s, 0.051s) \\
    \cline{2-3}
    & \multirow{3}{*}{Q'}
    & \propmethod & 
    -172 (-230, -130) & 0.106 (0.061, 0.19) & 0.241 (0.092, 1.1) & 0.964 (0.7, 1) & 1.94 (1.5, 2.4) & 0.946 (0.89, 0.97) & 1.2s (0.6s, 4.4s) \\
    && \mcmethod & 
    -172 (-230, -130) & - & 0.243 (0.091, 1.1) & 0.966 (0.74, 1) & 2.02 (1.6, 2.6) & 0.946 (0.89, 0.97) & 32s (24s, 38s) \\
    && \pgmethod & 
    -174 (-240, -130) & 2.59 (0.99, 5.3) & 0.252 (0.086, 1.1) & 0.81 (0.5, 0.99) & 1.33 (1.2, 1.4) & 0.946 (0.89, 0.97) & 0.12s (0.061s, 0.22s) \\
    \cline{2-3}
    & \multicolumn{2}{c|}{MCMC} & 
    - & - & 0.241 (0.091, 1.2) & 0.964 (0.71, 1) & 1.96 (1.5, 2.5) & 0.946 (0.89, 0.97) & 6m 41s (5m 42s, 7m 42s) \\
    \hline
    \multirow{7}{*}{500 / 25}
    & \multirow{3}{*}{Q}
    & \propmethod  & 
    -152 (-190, -120) & 0.00832 (0.0043, 0.015) & 1 (0.41, 3) & 0.912 (0.69, 0.99) & 3.36 (3, 4) & 0.98 (0.96, 0.99) & 4.5s (3.9s, 5.3s) \\
    && \mcmethod & 
    -151 (-190, -120) & - & 1 (0.42, 3) & 0.916 (0.69, 0.99) & 3.39 (3, 4) & 0.98 (0.96, 0.99) & 25s (23s, 29s) \\
    && \pgmethod & 
    -177 (-220, -130) & 5.52 (4, 8) & 1.11 (0.47, 3.8) & 0.732 (0.48, 0.91) & 2.34 (2.2, 2.6) & 0.98 (0.96, 0.99) & 0.043s (0.028s, 0.074s) \\
    \cline{2-3}
    & \multirow{3}{*}{Q'}
    & \propmethod & 
    -144 (-180, -110) & 0.406 (0.29, 0.55) & 0.996 (0.42, 3) & 0.938 (0.74, 0.99) & 3.57 (3.1, 4.3) & 0.98 (0.96, 0.99) & 1.8s (0.93s, 6.6s) \\
    && \mcmethod & 
    -144 (-180, -110) & - & 0.994 (0.42, 3) & 0.946 (0.77, 1) & 3.7 (3.2, 4.5) & 0.98 (0.96, 0.99) & 28s (24s, 32s) \\
    && \pgmethod & 
    -160 (-200, -120) & 8.41 (5.6, 13) & 1.12 (0.47, 3.8) & 0.722 (0.46, 0.9) & 2.29 (2.2, 2.5) & 0.98 (0.96, 0.99) & 0.18s (0.12s, 0.3s) \\
    \cline{2-3}
    & \multicolumn{2}{c|}{MCMC} & 
    - & - & 0.988 (0.42, 2.9) & 0.942 (0.77, 1) & 3.63 (3.1, 4.4) & 0.98 (0.96, 0.99) & 5m 23s (5m 18s, 5m 33s) \\
    \hline
    \multirow{7}{*}{1000 / 5}
    & \multirow{3}{*}{Q}
    & \propmethod  & 
    -407 (-550, -300) & 0.00135 (0.00045, 0.017) & 0.0444 (0.0091, 0.2) & 0.962 (0.63, 1) & 0.811 (0.63, 1.1) & 0.904 (0.8, 0.95) & 3.9s (3.6s, 4.6s) \\
    && \mcmethod & 
    -407 (-550, -300) & - & 0.0449 (0.0085, 0.21) & 0.96 (0.6, 1) & 0.811 (0.64, 1.1) & 0.904 (0.8, 0.95) & 1m 5s (37s, 1m 19s) \\
    && \pgmethod & 
    -430 (-600, -320) & 0.431 (0.1, 1.3) & 0.0427 (0.009, 0.2) & 0.869 (0.48, 1) & 0.618 (0.55, 0.7) & 0.904 (0.8, 0.95) & 0.014s (0.0067s, 0.026s) \\
    \cline{2-3}
    & \multirow{3}{*}{Q'}
    & \propmethod & 
    -406 (-550, -300) & 0.0417 (0.017, 0.085) & 0.0441 (0.0093, 0.2) & 0.989 (0.73, 1) & 0.877 (0.66, 1.2) & 0.904 (0.8, 0.95) & 0.86s (0.65s, 2.6s) \\
    && \mcmethod & 
    -406 (-550, -300) & - & 0.0437 (0.0092, 0.2) & 0.988 (0.72, 1) & 0.869 (0.65, 1.2) & 0.904 (0.8, 0.95) & 1m 1.3s (51s, 1m 9.1s) \\
    && \pgmethod & 
    -407 (-550, -300) & 0.716 (0.15, 1.8) & 0.042 (0.0088, 0.2) & 0.877 (0.49, 1) & 0.625 (0.56, 0.7) & 0.904 (0.8, 0.95) & 0.062s (0.029s, 0.12s) \\
    \cline{2-3}
    & \multicolumn{2}{c|}{MCMC} & 
    - & - & 0.0437 (0.0091, 0.2) & 0.989 (0.73, 1) & 0.863 (0.65, 1.2) & 0.904 (0.8, 0.95) & 5m 15s (5m 11s, 5m 22s) \\
    \hline
    \multirow{7}{*}{1000 / 10}
    & \multirow{3}{*}{Q}
    & \propmethod  & 
    -343 (-460, -250) & 0.00168 (0.00078, 0.0058) & 0.115 (0.04, 0.31) & 0.933 (0.76, 1) & 1.27 (1, 1.6) & 0.939 (0.88, 0.97) & 4.8s (4.3s, 6.2s) \\
    && \mcmethod & 
    -343 (-460, -250) & - & 0.115 (0.041, 0.32) & 0.935 (0.77, 1) & 1.27 (1.1, 1.6) & 0.939 (0.88, 0.97) & 1m 5.4s (54s, 1m 16s) \\
    && \pgmethod & 
    -388 (-530, -280) & 1.16 (0.51, 2.2) & 0.118 (0.042, 0.33) & 0.821 (0.58, 0.98) & 0.941 (0.86, 1.1) & 0.939 (0.88, 0.97) & 0.025s (0.013s, 0.049s) \\
    \cline{2-3}
    & \multirow{3}{*}{Q'}
    & \propmethod & 
    -341 (-460, -240) & 0.134 (0.069, 0.22) & 0.114 (0.04, 0.31) & 0.971 (0.84, 1) & 1.37 (1.1, 1.9) & 0.939 (0.88, 0.97) & 1.6s (0.84s, 4s) \\
    && \mcmethod & 
    -341 (-460, -240) & - & 0.114 (0.04, 0.32) & 0.969 (0.84, 1) & 1.36 (1.1, 1.9) & 0.939 (0.88, 0.97) & 1m 4.3s (54s, 1m 9.5s) \\
    && \pgmethod & 
    -345 (-470, -250) & 1.85 (0.7, 4.8) & 0.118 (0.043, 0.33) & 0.825 (0.59, 0.99) & 0.938 (0.85, 1.1) & 0.939 (0.88, 0.97) & 0.11s (0.058s, 0.22s) \\
    \cline{2-3}
    & \multicolumn{2}{c|}{MCMC} & 
    - & - & 0.115 (0.041, 0.31) & 0.973 (0.85, 1) & 1.37 (1.1, 1.9) & 0.939 (0.88, 0.97) & 5m 38s (5m 30s, 5m 49s) \\
    \hline
    \multirow{7}{*}{1000 / 25}
    & \multirow{3}{*}{Q}
    & \propmethod  & 
    -264 (-340, -230) & 0.00588 (0.0032, 0.01) & 0.493 (0.19, 1.5) & 0.906 (0.66, 0.99) & 2.44 (2.1, 2.7) & 0.977 (0.96, 0.98) & 12s (9.9s, 21s) \\
    && \mcmethod & 
    -264 (-340, -230) & - & 0.494 (0.2, 1.5) & 0.904 (0.65, 0.99) & 2.42 (2.1, 2.7) & 0.977 (0.96, 0.98) & 2m 24s (1m 56s, 2m 57s) \\
    && \pgmethod & 
    -332 (-440, -280) & 5.36 (3.6, 6.9) & 0.557 (0.21, 1.7) & 0.724 (0.47, 0.92) & 1.67 (1.5, 1.8) & 0.977 (0.96, 0.98) & 0.12s (0.054s, 0.37s) \\
    \cline{2-3}
    & \multirow{3}{*}{Q'}
    & \propmethod & 
    -256 (-330, -220) & 0.548 (0.42, 0.76) & 0.493 (0.2, 1.5) & 0.945 (0.72, 1) & 2.65 (2.2, 3.1) & 0.977 (0.96, 0.98) & 9.7s (3.8s, 28s) \\
    && \mcmethod & 
    -257 (-330, -220) & - & 0.491 (0.2, 1.5) & 0.949 (0.74, 1) & 2.66 (2.2, 3.1) & 0.977 (0.96, 0.98) & 2m 15s (1m 52s, 2m 49s) \\
    && \pgmethod & 
    -277 (-350, -240) & 8.15 (4.9, 12) & 0.554 (0.21, 1.7) & 0.723 (0.46, 0.93) & 1.66 (1.5, 1.8) & 0.977 (0.96, 0.98) & 0.57s (0.25s, 1.2s) \\
    \cline{2-3}
    & \multicolumn{2}{c|}{MCMC} & 
    - & - & 0.492 (0.2, 1.5) & 0.948 (0.74, 1) & 2.66 (2.2, 3.1) & 0.977 (0.96, 0.98) & 10m 46s (6m 49s, 14m 57s) \\
    \hline
    \multirow{7}{*}{10000 / 5}
    & \multirow{3}{*}{Q}
    & \propmethod  & 
    -4040 (-5600, -2700) & 0.0108 (0.0043, 0.1) & 0.00465 (0.00085, 0.027) & 0.948 (0.68, 1) & 0.259 (0.2, 0.37) & 0.898 (0.78, 0.96) & 48s (39s, 59s) \\
    && \mcmethod & 
    -4040 (-5600, -2700) & - & 0.00493 (0.00078, 0.027) & 0.951 (0.66, 1) & 0.254 (0.2, 0.36) & 0.898 (0.78, 0.96) & 14m 20s (4m 17s, 15m 28s) \\
    && \pgmethod & 
    -4350 (-6100, -2900) & 0.419 (0.068, 1.5) & 0.00495 (0.00074, 0.027) & 0.828 (0.45, 1) & 0.196 (0.17, 0.23) & 0.898 (0.78, 0.96) & 0.063s (0.023s, 0.23s) \\
    \cline{2-3}
    & \multirow{3}{*}{Q'}
    & \propmethod & 
    -4040 (-5600, -2700) & 0.0558 (0.019, 0.18) & 0.00489 (0.00083, 0.029) & 0.981 (0.76, 1) & 0.278 (0.2, 0.43) & 0.898 (0.78, 0.96) & 9.9s (6.5s, 25s) \\
    && \mcmethod & 
    -4040 (-5600, -2700) & - & 0.00494 (0.00075, 0.026) & 0.99 (0.77, 1) & 0.285 (0.21, 0.47) & 0.898 (0.78, 0.96) & 13m 5.7s (5m 16s, 14m 28s) \\
    && \pgmethod & 
    -4040 (-5600, -2700) & 0.842 (0.15, 3.9) & 0.00493 (0.00074, 0.027) & 0.836 (0.45, 1) & 0.199 (0.18, 0.23) & 0.898 (0.78, 0.96) & 0.31s (0.1s, 0.98s) \\
    \cline{2-3}
    & \multicolumn{2}{c|}{MCMC} & 
    - & - & 0.00568 (0.00074, 39) & 0.957 (0, 1) & 0.264 (0, 0.44) & 0.896 (0.76, 0.96) & 11m 7.8s (10m 24s, 12m 18s) \\
    \hline
    \multirow{7}{*}{10000 / 10}
    & \multirow{3}{*}{Q}
    & \propmethod  & 
    -3190 (-4300, -2100) & 0.00968 (0.0044, 0.045) & 0.0125 (0.0043, 0.038) & 0.93 (0.76, 1) & 0.414 (0.34, 0.54) & 0.939 (0.89, 0.97) & 40s (33s, 56s) \\
    && \mcmethod & 
    -3190 (-4300, -2100) & - & 0.0125 (0.0041, 0.039) & 0.926 (0.75, 1) & 0.412 (0.34, 0.53) & 0.939 (0.89, 0.97) & 13m 53s (6m 52s, 15m 29s) \\
    && \pgmethod & 
    -3740 (-5000, -2500) & 1.19 (0.54, 2.5) & 0.0125 (0.004, 0.041) & 0.807 (0.56, 0.98) & 0.301 (0.28, 0.35) & 0.939 (0.89, 0.97) & 0.13s (0.063s, 0.33s) \\
    \cline{2-3}
    & \multirow{3}{*}{Q'}
    & \propmethod & 
    -3190 (-4300, -2100) & 0.274 (0.12, 0.62) & 0.0126 (0.0044, 0.035) & 0.972 (0.87, 1) & 0.457 (0.35, 0.72) & 0.939 (0.89, 0.97) & 19s (8.9s, 46s) \\
    && \mcmethod & 
    -3190 (-4300, -2100) & - & 0.0128 (0.0041, 0.039) & 0.977 (0.85, 1) & 0.471 (0.37, 0.73) & 0.939 (0.89, 0.97) & 13m 3.3s (4m 24s, 14m 17s) \\
    && \pgmethod & 
    -3200 (-4300, -2100) & 2.73 (1.1, 8.4) & 0.0124 (0.0041, 0.041) & 0.817 (0.57, 0.99) & 0.304 (0.28, 0.35) & 0.939 (0.89, 0.97) & 0.72s (0.3s, 1.5s) \\
    \cline{2-3}
    & \multicolumn{2}{c|}{MCMC} & 
    - & - & 0.0127 (0.0041, 0.068) & 0.964 (0.74, 1) & 0.455 (0.34, 0.67) & 0.939 (0.89, 0.97) & 8m 53s (8m 40s, 9m 36s) \\
    \hline
    \multirow{7}{*}{10000 / 25}
    & \multirow{3}{*}{Q}
    & \propmethod  & 
    -2160 (-2900, -1700) & 0.0341 (0.0066, 0.13) & 0.0476 (0.025, 0.18) & 0.918 (0.65, 0.98) & 0.78 (0.67, 0.9) & 0.974 (0.95, 0.98) & 53s (34s, 1m 35s) \\
    && \mcmethod & 
    -2160 (-2900, -1700) & - & 0.0467 (0.026, 0.19) & 0.919 (0.65, 0.98) & 0.783 (0.67, 0.91) & 0.974 (0.95, 0.98) & 14m 15s (12m 49s, 15m 47s) \\
    && \pgmethod & 
    -3120 (-4100, -2400) & 4.89 (3.1, 7.6) & 0.0484 (0.026, 0.2) & 0.761 (0.46, 0.89) & 0.535 (0.49, 0.58) & 0.974 (0.95, 0.98) & 0.87s (0.48s, 1.6s) \\
    \cline{2-3}
    & \multirow{3}{*}{Q'}
    & \propmethod & 
    -2150 (-2900, -1700) & 1.72 (1, 3.9) & 0.0468 (0.026, 0.18) & 0.96 (0.78, 0.99) & 0.904 (0.73, 1.1) & 0.974 (0.95, 0.98) & 1m 4.1s (24s, 1m 50s) \\
    && \mcmethod & 
    -2160 (-2900, -1700) & - & 0.0475 (0.025, 0.18) & 0.971 (0.84, 1) & 0.958 (0.77, 1.2) & 0.974 (0.95, 0.98) & 13m 49s (10m 1.7s, 15m 33s) \\
    && \pgmethod & 
    -2170 (-2900, -1700) & 12.6 (7.5, 21) & 0.0483 (0.026, 0.2) & 0.764 (0.46, 0.9) & 0.539 (0.49, 0.58) & 0.974 (0.95, 0.98) & 3.9s (2.3s, 7.5s) \\
    \cline{2-3}
    & \multicolumn{2}{c|}{MCMC} & 
    - & - & 0.0469 (0.026, 0.19) & 0.959 (0.77, 0.99) & 0.89 (0.71, 1.1) & 0.974 (0.95, 0.98) & 18m 3.1s (12m 41s, 20m 44s) \\
    \hline
\end{tabular}%
}}
\end{table}
\egroup

\section{Computational Environment} \label{appendix:computational_environment}

The experiments were run on a server with the following specifications:

\subsection*{Hardware Information (Configuration 1)}
\begin{itemize}
  \item \textbf{CPU:} AMD EPYC 7742 64-Core Processor
  \item \textbf{CPU Cores:} 256
  \item \textbf{RAM:} 1.0Ti
\end{itemize}

\subsection*{Hardware Information (Configuration 2)}
\begin{itemize}
    \itemsep0em
    \item \textbf{CPU:} Intel(R) Xeon(R) CPU E5-2650 v4 @ 2.20GHz
    \item \textbf{CPU Cores:} 48
    \item \textbf{RAM:} 251Gi
\end{itemize}

\subsection*{Operating System Information}
\begin{verbatim}
  NAME="Red Hat Enterprise Linux"
  VERSION="8.5 (Ootpa)"
\end{verbatim}

Notably the logistic regression experiments in \Cref{sec:simulations} were run on Configuration 1, while the GP classification example and applications in \Cref{sec:application} were run on Configuration 2.

\subsection*{Software Information}

The software versions used for the experiments are as follows:
\begin{verbatim}
    python                    3.11.5
    pytorch                   2.1.0 
    gpytorch                  1.10  
    hamiltorch                0.4.1 
    torcheval                 0.0.7 
    numpy                     1.26.0
    matplotlib                3.7.2 
    geopandas                 0.14.1
    pandas                    2.1.3 
\end{verbatim}
Further information can be found in the \texttt{environment.yml} file in the supplementary material.

\end{document}